%% file: main.tex
\RequirePackage[svgnames]{xcolor}

\documentclass[11pt,logo,letterpaper]{berkeley}

\usepackage[utf8]{inputenc} %
\usepackage[T1]{fontenc}    %
\usepackage{lmodern}
\usepackage{textcomp}
\usepackage[svgnames]{xcolor}
\usepackage{asymptote}
\usepackage[comma,authoryear,compress]{natbib}
\usepackage{hyperref}
\usepackage{algorithm}
\usepackage{algorithmicx}
\usepackage{algpseudocode}
\usepackage{microtype}
\usepackage{graphicx}
\expandafter\def\csname ver@subfig.sty\endcsname{}
\usepackage{booktabs} %
\usepackage{float}
\usepackage{bigstrut}
\usepackage{tikz}
\usetikzlibrary{tikzmark}
\makeatletter
\newcommand*\myfontsize{%
  \@setfontsize\myfontsize{7}{8}%
}
\makeatother
\definecolor{myred}{rgb}{0.7, 0.3, 0.0}
\definecolor{myblue}{HTML}{054488}
\definecolor{mygreen}{HTML}{056b34}

\newcolumntype{R}[1]{>{\raggedleft\let\newline\\\arraybackslash\hspace{0pt}}m{#1}}

\usetikzlibrary{intersections}

\definecolor{darkgreen}{rgb}{0.0, 0.42, 0.24}

\usepackage{amsthm}
\usepackage{nicefrac}
\usepackage{subcaption}
\usepackage{caption}
\usepackage{graphicx}
\usepackage{cleveref}
\usepackage{bxcoloremoji}
\usepackage{listings}
\usepackage{float}
\usepackage{longtable}
\lstdefinestyle{python}{
    language=Python,
    basicstyle=\ttfamily\footnotesize,
    keywordstyle=\color{blue}\bfseries,
    commentstyle=\color{green},
    stringstyle=\color{red},
    numberstyle=\tiny\color{gray},
    showstringspaces=false,
    frame=single,
    breaklines=true,
    backgroundcolor=\color{lightgray!20}
}

\usepackage{hyperref}       %
\usepackage{url}            %
\usepackage{booktabs}       %
\usepackage{nicefrac}       %
\usepackage{microtype}      %
\usepackage{graphicx}
\usepackage{subcaption} 
\usepackage{wrapfig}
\usepackage{lipsum}
\usepackage{enumitem}
\usepackage{stackengine}
\usepackage[font=small,labelfont=bf]{caption}
\usepackage{color}
\usepackage{adjustbox}
\usepackage{tikz}
\usepackage{tcolorbox}
\usepackage{rotating}
\usepackage{makecell}
\usepackage{colortbl}    
\usepackage{multirow}    
\usepackage{pifont}      
\usepackage{xcolor}      
\usepackage{array}       
\usepackage{amsmath}
\usepackage{amsfonts}       %
\usepackage{xspace}
\usepackage{mathtools}
\definecolor{oursblue}{RGB}{230,240,255} 

\input{macro}

\newtcolorbox{AIbox}[2][]{aibox,title=#2,#1}
\definecolor{lightblue}{rgb}{0.22,0.45,0.70}%
\definecolor{Gray}{gray}{0.95}
\definecolor{Cornsilk}{rgb}{1.0, 0.97, 0.86}
\usepackage{enumitem}

\makeatother

\definecolor{myred}{rgb}{0.7, 0.3, 0.0}
\definecolor{myblue}{HTML}{054488}
\definecolor{mygreen}{HTML}{056b34}
\definecolor{myorange}{HTML}{ff8800}
\definecolor{mypurple}{HTML}{8400ff}
\definecolor{mypink}{HTML}{f7acb9}

\definecolor{myred}{rgb}{0.7, 0.3, 0.0}
\definecolor{myblue}{HTML}{054488}
\definecolor{mygreen}{HTML}{056b34}
\definecolor{tiktokpink}{HTML}{E91E63}
\definecolor{tiktokpurple}{HTML}{673AB7}
\definecolor{tiktokgray}{HTML}{9E9E9E}

\newcommand{\mytitle}{ST-ResGAT: Explainable Spatio-Temporal Graph Neural Network for Road Condition Prediction and Priority-Driven Maintenance}

\usepackage{amssymb}

\title{\mytitle}

\author{
Mohsin Mahmud Topu$^{1}$,
Azmine Toushik Wasi$^{1}$,
Mahfuz Ahmed Anik$^{1}$,
MD Manjurul Ahsan$^{2}$
}

\affil{
$^1$Shahjalal University of Science and Technology, Sylhet, Bangladesh\\
\vspace{-2.5mm}
$^2$University of Oklahoma, Norman, OK, United States
}

\begin{document}

\begin{abstract}
\textbf{Abstract:} Climate-vulnerable road networks require a paradigm shift from reactive, fix-on-failure repairs to predictive, decision-ready maintenance. This paper introduces ST-ResGAT, a novel Spatio-Temporal Residual Graph Attention Network that fuses residual graph-attention encoding with GRU temporal aggregation to forecast pavement deterioration. Engineered for resource-constrained deployment, the framework translates continuous Pavement Condition Index (PCI) forecasts directly into the American Society for Testing and Materials (ASTM)-compliant maintenance priorities. Using a real-world inspection dataset of 750 segments in Sylhet, Bangladesh (2021–2024), ST-ResGAT significantly outperforms traditional non-spatial machine learning baselines, achieving exceptional predictive fidelity ($R^2 = 0.93$, $RMSE = 2.72$). Crucially, ablation testing confirmed the mathematical necessity of modeling topological neighbor effects, proving that structural decay acts as a spatial contagion. Uniquely, we integrate GNNExplainer to unbox the model, demonstrating that its learned priorities align perfectly with established physical engineering theory. Furthermore, we quantify classification safety: achieving $85.5\%$ exact ASTM class agreement and $100\%$ adjacent-class containment, ensuring bounded, engineer-safe predictions. To connect model outputs to policy, we generate localized longitudinal maintenance profiles, perform climate stress-testing, and derive Pareto sustainability frontiers. 
ST-ResGAT therefore offers a practical, explainable, and sustainable blueprint for intelligent infrastructure management in high-risk, low-resource geological settings.

\vspace{0.5cm}

\coloremojicode{1F4C5} \textbf{Date}: March 15, 2026

\coloremojicode{1F4E7} \textbf{Correspondence}: Mohsin Mahmud Topu~(\href{mailto:mohsinmahmudtopu@gmail.com}{mohsinmahmudtopu@gmail.com})

\coloremojicode{1F4BB} \textbf{Keywords}: Pavement Condition Index, Predictive Maintenance,\\ Explainable AI, Climate Resilience, Graph Attention Networks


\end{abstract}

\maketitle

\begin{figure}[h]
    \centering
    \includegraphics[width=\linewidth]{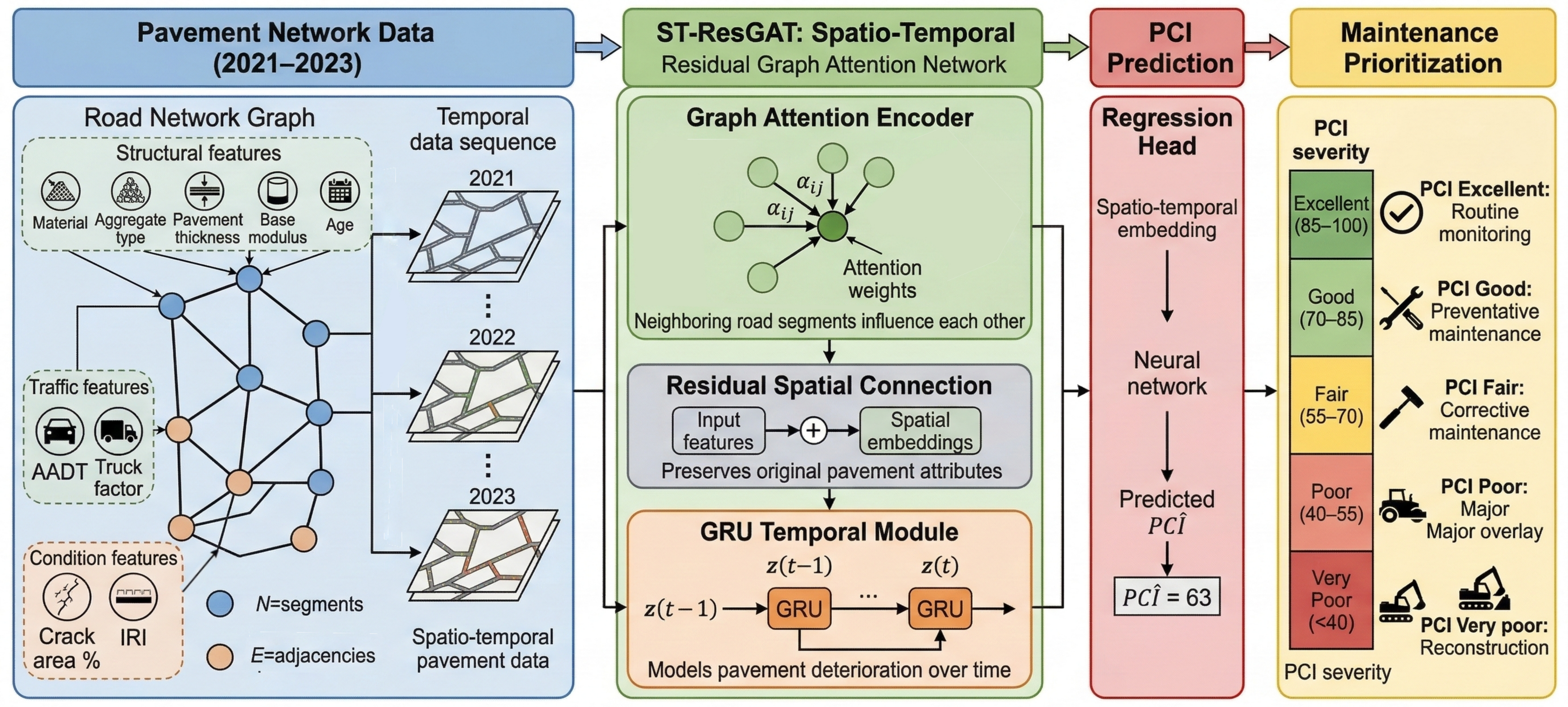}
    \caption{\textbf{Graphical Abstract.} Overview of ST-ResGAT development and evaluation.}
\end{figure}

\section{Introduction}\label{sec1}
Road infrastructure is a foundational component of modern transportation systems, enabling economic mobility, emergency response, and regional connectivity \citep{lestari2025interplay}. Yet the maintenance of large-scale pavement networks remains a persistent challenge for transportation agencies worldwide. Most road authorities continue to rely on reactive maintenance strategies, where interventions occur only after severe deterioration has already manifested \citep{famewo2025review}. This \textit{fix-on-failure} paradigm accelerates structural degradation, increases vehicle operating costs, and introduces safety risks across transportation networks \citep{samuel2003effect}. It also produces substantial environmental consequences because repeated rehabilitation cycles consume large volumes of aggregates, asphalt binders, and energy-intensive construction resources, contributing to the growing carbon footprint of infrastructure systems \citep{mosly2025carbon}. Transitioning from reactive management to predictive, data-driven maintenance planning has therefore become a central objective in sustainable infrastructure management and in achieving several targets within the United Nations Sustainable Development Goals (SDGs).

Recent advances in machine learning and deep learning have accelerated the development of automated pavement condition assessment and prediction frameworks \citep{berangi2025structural}. Early studies employed classical algorithms such as Support Vector Machines (SVM) \citep{basavaraju2019machine} and Artificial Neural Networks (ANN) \citep{kheirati2022machine} to estimate pavement condition indices from inspection data. While these approaches demonstrated the feasibility of data-driven condition modeling, they often struggled to capture the complex interactions among environmental, structural, and operational factors that influence pavement deterioration \citep{chen2023review}. More recent research has explored Convolutional Neural Networks (CNN) for image-based distress detection \citep{das2021application, li2019image}, Graph Neural Networks (GNN) for representing road network topology \citep{gao2024considering}, and Digital Twin (DT) systems for high-resolution infrastructure monitoring \citep{sierra2022development, yan2023digital}. Hybrid approaches combining GNN and DT frameworks have also been proposed for predictive maintenance applications \citep{lu2025modeling,topu2025digital}. Despite these advances, practical deployment remains limited due to data requirements, infrastructure costs, and computational complexity.

A critical methodological limitation in the existing literature is that many predictive models treat pavement segments as independent entities, ignoring the inherent spatial interdependence of road networks. In practice, deterioration processes propagate across connected segments through shared loading patterns, drainage conditions, and environmental exposure \citep{sadeghian2025pixels}. Ignoring these network dependencies reduces the ability of models to capture realistic degradation dynamics. At the same time, advanced digital monitoring frameworks frequently depend on high-quality automated inspection systems whose accuracy typically ranges between 85–90\%, still below manual surveys \citep{luo2022improving,pierce2013practical}. These constraints are particularly problematic for transportation agencies operating under limited data availability and resource constraints. Furthermore, even when predictive models are developed, they rarely integrate with operational maintenance planning. Existing scheduling approaches often rely on retrospective data and optimization techniques such as Genetic Algorithms \citep{chiou2025study} or multi-criteria decision-making methods \citep{sayadinia2021proposing}, which remain disconnected from forward-looking deterioration predictions. Consequently, a gap persists between predictive analytics and actionable maintenance prioritization.

To address these challenges, this study proposes ST-ResGAT, an explainable spatio-temporal residual graph attention network designed to model pavement deterioration across interconnected road networks while simultaneously supporting maintenance prioritization. The proposed architecture integrates graph attention mechanisms to capture spatial dependencies, residual connections to stabilize deep graph learning, and gated temporal aggregation to represent longitudinal deterioration dynamics. Unlike pointwise regression models or conventional temporal predictors, ST-ResGAT explicitly incorporates network topology and neighbor influence when forecasting future Pavement Condition Index (PCI) trajectories. The framework further bridges the gap between prediction and operational decision-making by translating continuous PCI forecasts into ASTM D6433 condition categories \citep{ASTM-D6433-23} and generating priority-driven maintenance profiles.
The model is evaluated using a longitudinal pavement inspection dataset containing 3,000 observations across 750 road segments collected between 2021 and 2024. Experimental results demonstrate that ST-ResGAT consistently outperforms five baseline methods, achieving an $R^2$ of 0.93 on held-out data. To enhance transparency and practitioner trust, explainable-AI techniques based on GNNExplainer \citep{ying2019gnnexplainer} are adapted to the spatio-temporal graph learning setting, enabling both local and global interpretation of deterioration drivers. The framework additionally evaluates classification safety by examining the alignment between predicted PCI categories and ASTM thresholds, thereby quantifying the risk of maintenance misclassification in practical deployment.

The main contributions of this paper are:

\begin{enumerate}
    \item We develop \textbf{ST-ResGAT}, an explainable spatio-temporal residual graph attention network that integrates graph attention mechanisms, residual learning, and gated temporal aggregation to model spatial dependencies and temporal deterioration dynamics in road networks.
    
    \item We introduce a \textbf{predictive-to-decision framework} that converts continuous Pavement Condition Index (PCI) forecasts into ASTM D6433-compliant condition categories \citep{ASTM-D6433-23} and generates segment-level maintenance priority rankings to support practical asset management.
    
    \item We extend \textbf{explainable artificial intelligence techniques for spatio-temporal GNNs}, adapting feature-attribution and perturbation-based analyses to identify dominant deterioration drivers and improve interpretability for infrastructure engineers.
    
    \item We conduct a detailed \textbf{empirical evaluation} using a longitudinal inspection dataset covering 750 pavement segments from 2021–2024, demonstrating improved predictive accuracy, robust ASTM category alignment, and operationally actionable maintenance prioritization.
\end{enumerate}

\begin{table}[htbp]
\centering
\caption{Abbreviations and Acronyms}
\label{tab:abbreviations}
\begin{tabular}{p{2cm} p{15cm}}
\hline
\textbf{Abbreviation} & \textbf{Full Form} \\
\hline
AADT & Annual Average Daily Traffic \\
AdaBoost & Adaptive Boosting \\
AI & Artificial Intelligence \\
ANN & Artificial Neural Networks \\
ASTM & American Society for Testing and Materials \\
CatBoost & Categorical Boosting \\
CDV & Corrected Deduct Value \\
CNN & Convolutional Neural Networks \\
DL & Deep learning \\
DT & Digital Twin \\
EPT & Effective Pavement Thickness \\
FE & Finite-element \\
GA & Genetic Algorithms \\
GAT & Graph Attention Network \\
GNN & Graph Neural Network \\
GPR & Ground Penetrating Radar \\
GRU & Gated Recurrent Unit \\
IRI & International Roughness Index \\
LIDAR & Light Detection and Ranging \\
LSTM & Long Short-Term Memory \\
LTPP & Long-Term Pavement Performance \\
MAE & Mean Absolute Error \\
ML & Machine Learning \\
MLP & Multilayer Perceptron \\
MSE & Mean Squared Error \\
PCI & Pavement Condition Index \\
REC & Regression Error Curve \\
RF & Random Forest \\
RHD & Roads and Highway Department \\
RL & Reinforcement Learning \\
RMMS & Road Maintenance Management System \\
RMSE & Root Mean Squared Error \\
RUL & Remaining Useful Life \\
SDGs & Sustainable Development Goals \\
SNO & Simultaneous Network Optimization \\
ST-GNNs & Spatio-temporal Graph Neural Networks \\
ST-ResGAT & Spatio-Temporal Residual Graph Attention Network \\
STGAT & Spatial-Temporal Graph Attention Network \\
SVM & Support Vector Machines \\
UAV & Unmanned Aerial Vehicle \\
XAI & Explainable Artificial Intelligence \\
XGBoost & Extreme Gradient Boosting \\
\hline
\end{tabular}
\end{table}

The remainder of the paper is organized as follows. Section ~\ref{sec2} reviews related work on pavement health prediction, graph neural networks for infrastructure, and priority-based maintenance in road health monitoring. Section ~\ref{sec3} describes the problem formulation and methodology, and details the ST-ResGAT model and the predictive-to-decision translation. Section 4 describes the dataset and experimental setup. Section ~\ref{sec5} and ~\ref{sec6} reports experimental results, model diagnostics, and ablation studies. Section ~\ref{sec7} discusses deployment considerations, SDG goal alignment, practical implication of the study. Section ~\ref{sec8} details the limitations and future scopes of the study and Section ~\ref{sec9} concludes with the findings summary of the study. Table \ref{tab:abbreviations} lists all the abbreviations and acronyms used in this paper.

\section{Related Works}\label{sec2}

This section reviews prior research pavement health prediction, graph-based models in pavement deterioration modeling, and the recent advancements in priority-based maintenance optimization frameworks are reviewed. We discuss established approaches to road damage detection and predictive maintenance. Finally, we identify methodological gaps that motivate the development of the proposed predictive maintenance framework.

\subsection{ML-based Road Condition Prediction}

The quest to automate the evaluation of the PCI has transitioned from traditional empirical equations to robust machine learning (ML) architectures \citep{yuan2024automated,ani2025machine,daghigh2024review}. Leveraging large observational repositories such as the Long-Term Pavement Performance (LTPP) dataset, researchers have demonstrated the effectiveness of ensemble learners: for instance, \cite{piryonesi2021examining} trained Random Forest and gradient-boosting classifiers on LTPP and reported categorical prediction accuracies above 85\%. Similarly, XGBoost has been applied to highway concrete distress prediction \citep{lee2020predicting}, while comparative studies have evaluated naive Bayes, boosted forests, $k$-nearest neighbors and multivariable linear regression for crack-rating tasks \citep{inkoom2019pavement}. A growing body of work confirms that ML integration materially enhances pavement-damage forecasting accuracy \citep{wu2020automated,nabipour2019comparative,basavaraju2019machine,ani2025machine}.

Concurrently, deep learning (DL) approaches have gained traction because of their capacity to learn complex, nonlinear relationships and hierarchical feature representations \citep{peng2024asphalt}. Feedforward ANNs provide flexible function approximation but remain sensitive to hyperparameterization \citep{radwan2025comparative,karballaeezadeh2020smart}; convolutional architectures excel at capturing spatial patterns relevant to PCI prediction \citep{majidifard2020deep}; and recurrent models, notably LSTMs, are commonly adopted for temporal dynamics \citep{gowda2025optimizing, choi2019development}. Despite these advances, two practical gaps persist: most models either neglect or inadequately represent spatial neighbor-effects that govern damage propagation, and they seldom translate continuous forecasts into engineer-relevant condition bands with quantified safety margins. These limitations hinder the readiness of ML outputs for operational decision-making by practitioners.

\subsection{Graph-based Models for Road Health Prediction}

GNNs conceptualize the highway network as a non-Euclidean system of interconnected nodes and edges. This shift acknowledges that pavement deterioration is a topologically dependent process, where structural distress in one segment inevitably influences the degradation rate of adjacent sections \citep{tong2025stgan}. \cite{gao2024considering} demonstrated that by utilizing GraphSAGE, models could successfully capture these spatial correlations, improving the accuracy of road  showing improvements in $R^2$ scores ranging from 0\% to 20\% over traditional machine learning regressors. The evolution of this field continued with the introduction of \textit{attention} mechanisms, which allow for a more nuanced understanding of network influence \citep{wasi2024graph}.

Spatio-temporal Graph Neural Networks (ST-GNNs) have emerged as the leading paradigm for forecasting over networked infrastructure because they jointly model topology and time \citep{corradini2025systematic}. Contemporary architectures combine a learned spatial encoder (graph convolutions or attention) with a temporal module (RNNs, temporal convolutions, or attention) to capture how node states evolve under neighbor influence \citep{rahmani2023graph,guo2023capturing}. A comparative summary of representative spatio-temporal architectures, their limitations for pavement condition forecasting, and the targeted experiments used in this study is presented in Table~\ref{tab1}.

\begin{table*}[htbp]
\centering
\caption{Graph-based infrastructure prediction models: overview, research gaps, and how ST-ResGAT addresses them.}
\label{tab1}
\resizebox{\linewidth}{!}{
\begin{tabular}{p{3cm} p{5.5cm} p{5.5cm} p{5.5cm}}
\toprule
\textbf{Author / Year (Model)} & \textbf{Application} & \textbf{Gaps} & \textbf{ST-ResGAT Contribution} \\
\midrule

\cite{gao2024considering} (GraphSAGE) &
GNN applied on road-network topology to model pavement deterioration using PMIS data. &
Neighbour effects modeled implicitly; limited interpretability for infrastructure decisions. &
Graph attention with residual projection quantifies neighbour influence and provides explainable spatial learning. \\

\addlinespace

\cite{zhou2024graph} (FE-GNN surrogate) &
Graph neural surrogate trained on 3D finite-element pavement response simulations. &
Focuses on structural response prediction rather than network-level PCI forecasting. &
Extends graph learning toward segment-level PCI prediction with temporal modeling. \\

\addlinespace

\cite{tong2025stgan} (STGAN) &
Spatio-temporal graph autoregression combining graph attention and temporal autoregressive prediction. &
Feature mixing during message passing; limited interpretability for practitioners. &
Residual spatial representation stabilizes feature propagation with explainable node influence. \\

\addlinespace

\cite{dang2022structural} (g-SDDL) &
GNN + convolutional stack operating directly on raw vibration signals; stacked models for multi-damage detection without hand-engineered features. &
Focused on classification/damage detection from vibration data; not designed for network-level continuous PCI forecasting or long-horizon deterioration. &
Inspires raw-sensor→node feature pipeline in ST-ResGAT (use raw time-series as node inputs) and ensemble/stacking strategies for robust multi-damage segment modelling while preserving ST-ResGAT’s temporal forecasting and explainability. \\

\addlinespace

\cite{djenouri2022intelligent} (Intelligent GCN crack detection) &
Image-to-graph conversion with GCN for road crack detection. &
Patch-level detection; ignores network topology and temporal deterioration. &
Segment-level graph representation capturing spatial propagation of pavement conditions. \\

\addlinespace

\cite{feng2023scl} (SCL-GCN) &
Contrastive-learning enhanced GCN for LiDAR-based crack detection. &
Detection-focused; lacks temporal forecasting and maintenance decision mapping. &
Combines spatial GAT with temporal GRU to enable deterioration forecasting. \\

\addlinespace

\cite{liu2025graph} (GPS-GNN Pavement Simulator) &
Encoder–Processor–Decoder GNN surrogate trained on 3D FE pavement simulations. &
High-fidelity FE surrogate but not designed for decision-ready PCI forecasting. &
Leverages structural response patterns and maps outputs to maintenance condition bands. \\

\addlinespace

\cite{su2026physics} (GNN–Transformer multitask) &
Physics-guided GNN models microstructure topology while Transformer analyzes stochastic load spectra for fatigue prediction. &
Microstructure-scale modeling; limited applicability to road-network level deterioration. &
ST-ResGAT focuses on segment-level spatial topology and temporal deterioration for network-scale forecasting. \\

\addlinespace

\cite{he2024synthesizing} (HeteroGNN + ontology) &
Heterogeneous GNN with bridge defect ontology to predict preservation activities. &
Focused on bridge maintenance classification rather than condition forecasting. &
Adopts explainable graph reasoning to connect predicted PCI with maintenance prioritization. \\

\addlinespace

\cite{kong2024spatio} (STP-GNN) &
Extended message-passing GNN modeling spatio-temporal degradation propagation for remaining useful life prediction. &
Designed for equipment RUL; limited infrastructure-specific interpretability. &
Residual spatial attention and explainability tailored for pavement network deterioration modeling. \\

\bottomrule
\end{tabular}}
\end{table*}

\subsection{Maintenance Optimization and Scheduling}

Historically, the optimization of maintenance schedules has been addressed as a constrained mathematical problem, primarily utilizing metaheuristic algorithms to balance budget limitations with network performance \citep{wettewa2024graph, li2025advancing}. Early models typically balanced limited budgets against network performance goals using metaheuristic solvers. For example, \cite{yamany2024network} utilized Genetic Algorithms (GA) to optimize multi-year budget allocations for highway agencies, demonstrating that automated scheduling could improve overall network health compared to manual prioritization. Similarly, \cite{zhu2025novel} developed a novel pavement maintenance decision model integrating crack causes to enhance decision accuracy and maintenance efficacy. \cite{yao2022large} constructed an RL simulation for a multi-lane highway network and achieved roughly 26.6\% lifecycle cost savings compared to a traditional threshold-based scheme. Extending this idea, \cite{yao2024multi} formulated a multi-agent RL model that explicitly captures the interdependence of adjacent segments. In a real-world highway network case, their “simultaneous network optimization” (SNO) approach produced about 3.0\% total cost reduction and up to 17.5\% improvement in average pavement performance. These studies demonstrate that data-driven optimization can significantly outperform manual budgeting.

In parallel, the research community is converging on all‑in‑one “digital twin” frameworks that integrate damage detection with scheduling. A key trend is to embed graph-based predictive models within a real-time monitoring platform. For instance, \cite{topu2025digital} propose a pavement digital twin where UAV, LiDAR, and embedded sensor streams continuously update a graph model of the road network. A graph neural network (GNN) learns from these spatiotemporal inputs – including physical attributes, traffic loads, and environment data – to forecast segment deterioration. Similarly, \cite{lu2025modeling} developed a Spatial–Temporal Graph Attention network (STGAT) within a highway digital twin. By fusing heterogeneous past and real-time data (e.g. roughness, cracking, traffic volume), the STGAT accurately predicts future pavement conditions. Furthermore, hybrid frameworks combining Fuzzy Logic with GNN-derived predictions have been investigated to handle the uncertainties inherent in environmental stressors and material aging \citep{santos2022fuzzy}, providing a more resilient decision-support layer for asset managers.

\subsection{Research Gap Analysis}

While machine learning and graph-based approaches have demonstrated promising performance in road damage prediction \citep{gao2024considering, kong2024spatio, yuan2024automated, ani2025machine}, the majority of existing studies remain confined to only damage prediction or classification. These works primarily determine the risk of damage at some point in future, but do not formalize the subsequent decision process required to identify which specific road/segment should be treated first. Even in advanced ST-GNN frameworks originally designed for traffic forecasting \citep{bui2022spatial, zhong2025stte}, model formulations are typically guided by temporal patterns that are smooth or periodic rather than by mechanisms that explicitly capture abrupt, event-driven deterioration such as the rapid PCI loss observed after severe flooding \citep{masuda2016modelling}. Current all-in-one solutions typically rely on DT architectures \citep{lu2025modeling}, which requires prohibitive capital investment and specialized technical expertise, which is not feasible in resource-constrained low-income countries. Furthermore, explainability methods such as GNNExplainer, commonly applied to graphs remain largely unexplored in this field. As a consequence, prediction, explanation, and intervention design are often treated as sequential stages rather than as a structurally integrated framework. Consequently, there is a methodological scarcity of frameworks tailored for hyper-vulnerable, monsoon-driven environments like Sylhet, Bangladesh. In such climate-risk–intensive contexts, infrastructure management continues to rely largely on reactive maintenance practices, which not only result in substantial resource inefficiencies but also exacerbate public disruption, environmental degradation, and accident risks. These conditions underscore the urgent need for a feasible and context-aware framework capable of supporting proactive infrastructure management. Yet, within road health monitoring research, the methodological integration of spatio-temporal forecasting with a robust maintenance profiling and prioritization layer, especially in low-resource geological settings remain limited.

To address these limitations, we proposed a spatio-temporal residual graph attention framework that connects prediction, explanation, and priority-based ranking for efficient maintenance within a unified architecture. Rather than treating interpretability and prediction accuracy as terminal analytical outputs, model explanations derived through GNNExplainer are integrated into the decision framework to ensure that maintenance recommendations remain both physically interpretable and operationally reliable. Furthermore, the reliability of AI-driven recommendations is seldom evaluated through a risk-management perspective, as current frameworks largely overlook classification safety and marginal error analysis, leaving boundary-case behaviour insufficiently examined despite its potential implications for catastrophic infrastructure failures.

Beyond methodological robustness, these challenges also carry broader sustainability implications in countries like Bangladesh, where progress toward the UN SDG goals continues to face structural pressures stemming from climate vulnerability, infrastructure resilience, and sustainable urban mobility demands. By integrating interpretable spatio-temporal forecasting with maintenance prioritization and risk-aware evaluation, the proposed framework implicitly supports broader sustainability objectives including economic productivity, resilient cities, responsible resource use, and climate action. In doing so, the study moves beyond descriptive infrastructure modelling toward a sustainability-oriented decision paradigm in which predictive insights are translated into operational strategies that support measurable progress toward global development targets while enabling resilient highway asset management under intensifying climate stress.

\section{Motivation}
Road networks in climate-vulnerable regions experience accelerated deterioration due to extreme hydrological and environmental conditions. Bangladesh represents a particularly relevant context because a large share of its population and transportation infrastructure is exposed to seasonal flooding, high precipitation variability, and rapid urban expansion. Approximately 56\% of the national population resides in areas with high exposure to climate-related hazards \citep{tawfique2024bangladesh}. These environmental stresses significantly influence pavement performance by increasing moisture infiltration, weakening subgrade layers, and intensifying load-induced fatigue processes.

The northeastern Sylhet division illustrates the severity of these challenges. During the 2022 flood event, approximately 55.76\% of the region was inundated, affecting nearly 10.6 million residents and submerging roughly 43.38\% of major roadways \citep{shafiq2023frequency}. Flood-induced damage accelerates pavement distress mechanisms such as stripping, pothole formation, and structural base failure, resulting in rapid declines in Pavement Condition Index (PCI). The resulting disruptions extend beyond engineering concerns by increasing transportation costs, interrupting freight movement, and limiting access to essential services.

These environmental shocks rarely affect road segments in isolation. Floodwater propagation, drainage connectivity, and traffic redistribution often produce spatially correlated deterioration patterns across adjacent segments of a transportation network. Consequently, pavement degradation in one location may influence the performance of neighboring links through altered loading patterns, shared environmental exposure, and maintenance deferral effects. Such network-level dependencies highlight the need for analytical frameworks capable of modeling spatial interactions alongside temporal deterioration dynamics.

At the same time, transportation agencies in many developing regions operate under constraints including limited inspection data, restricted maintenance budgets, and reduced access to high-cost monitoring technologies such as Digital Twin systems. These constraints limit the feasibility of infrastructure monitoring approaches that rely on dense sensor networks or continuous high-resolution surveys. As a result, there is significant value in developing predictive models that are both data-efficient and operationally interpretable, enabling proactive maintenance planning within resource-constrained transportation management environments.

\begin{figure*}[t]
    \centering
    \includegraphics[width=\linewidth]{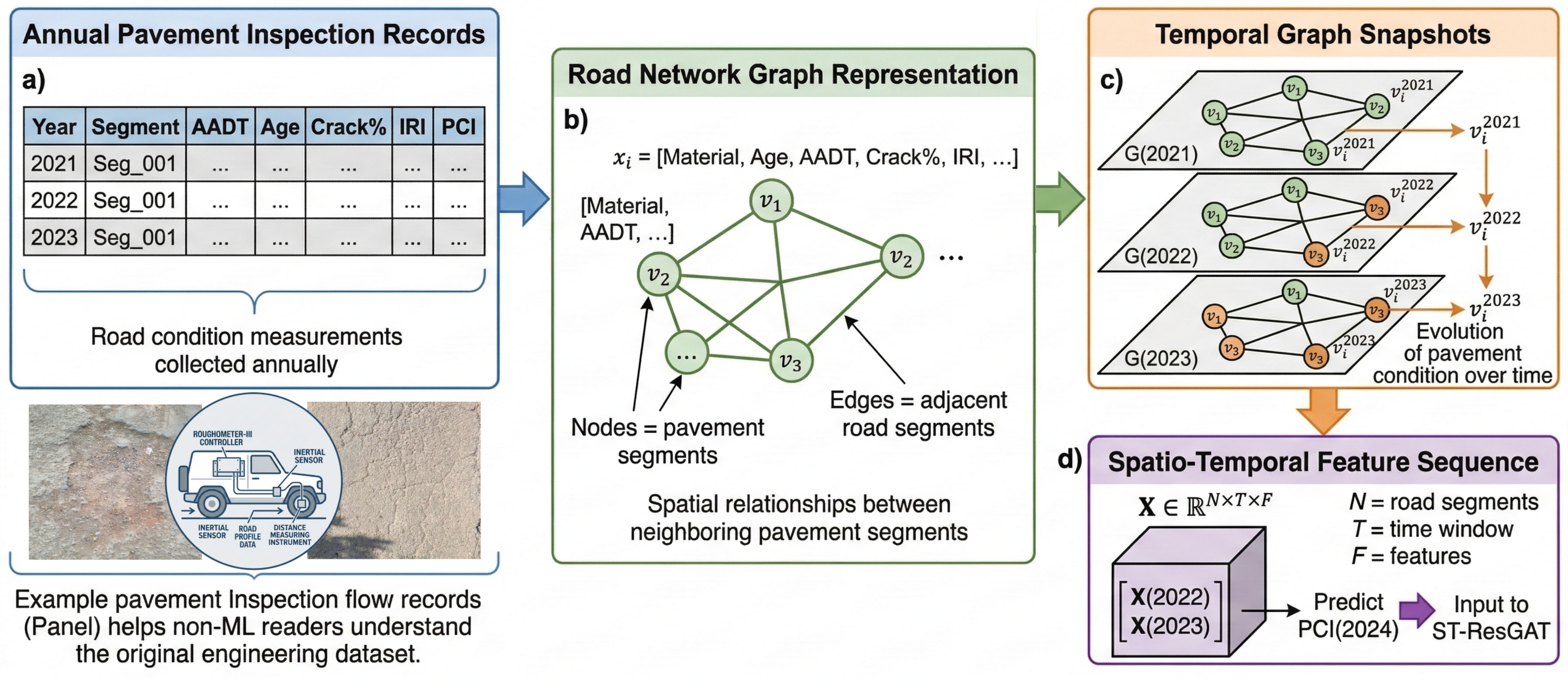}
    \caption{\textbf{Construction of the spatio-temporal graph representation used in ST-ResGAT}. Pavement inspection records collected across multiple years are first organized as node features for each road segment. THe road network topology defines adjacency relationships among segments, forming a graph structure. Each year corresponds to a graph snapshot with identical topology but updated node attributes. A temporal window of historical snapshots is then combined to create a spatio-temporal feature sequence that serves as input to the proposed ST-ResGAT model for future pavement condition prediction.}
    \label{fig2}
\end{figure*}

\section{Methodology}\label{sec3}
This section details the formulation, architecture, and validation strategy of the Spatio-Temporal Residual Graph Attention Network (ST-ResGAT). To accurately capture the complex, interdependent dynamics of pavement deterioration, the physical road infrastructure is first mathematically formulated as a non-Euclidean graph, integrating multi-dimensional structural, traffic, and condition attributes. The proposed deep learning architecture synergistically couples a multi-head Graph Attention mechanism, enhanced with residual connections to capture the spatial contagion of structural damage with a Gated Recurrent Unit (GRU) to model sequential temporal decay. Furthermore, to ensure the framework's viability as an operational decision-support tool, this section outlines the integration of explainable AI (GNNExplainer) for mechanistic transparency, alongside the downstream translation of continuous Pavement Condition Index (PCI) forecasts into standardized, engineer-safe ASTM maintenance categories.
\subsection{Problem Formulation}
Consider a road network consisting of $N$ pavement segments observed over $T$ years. Each segment is represented as a node in a graph, and adjacency relationships between road segments define graph edges. The objective is to predict the \textit{Pavement Condition Index (PCI)} of each segment for a future year using structural, traffic, and historical attributes. Table \ref{tab:notations} lists all the notations used.

Formally, the road network is represented as a graph
$G = (V, E),$
where $V = \{v_1, v_2, ..., v_N\}$ denotes the set of pavement segments and $E$ represents adjacency relationships between segments.
For each node $v_i$ at time $t$, we define a feature vector
$\mathbf{x}_i^{(t)} \in \mathbb{R}^{F},$
where $F$ denotes the number of attributes. The features are grouped into three categories in order to capture different aspects of pavement deterioration and loading behavior. The first group consists of \textit{structural features} (i) including pavement material type, aggregate type, effective pavement thickness (EPT), base modulus, and pavement age, which describe the physical and mechanical properties of the pavement structure. The second group includes \textit{traffic features} (ii) such as Annual Average Daily Traffic (AADT) and truck factor, representing vehicular loading intensity and heavy vehicle impact on pavement performance \citep{kumar2025effects}. The third group contains \textit{condition features} (iii) including crack area percentage and the International Roughness Index (IRI), which characterize the current surface distress and ride quality of the pavement \citep{paterson1986international}.

The prediction target is the PCI value
$y_i^{(t)} \in \mathbb{R}.$
Given historical observations for $T_0$ previous years, the goal is to estimate
\begin{equation}
\hat{y}_i^{(t+1)} = f_{\theta}\left(
\mathbf{x}_i^{(t-T_0+1)}, \ldots, \mathbf{x}_i^{(t)}, G
\right),
\end{equation}
where $f_{\theta}(\cdot)$ denotes the proposed \textbf{Spatio Temporal Residual Graph Attention Network (ST-ResGAT)} parameterized by $\theta$.
The predicted PCI values are later used for maintenance prioritization through \cite{ASTM-D6433-23} based severity categorization, although the prioritization procedure itself is independent of the graph model.

\subsection{Data Pre-processing and Graph Construction}
\noindent \textbf{Dataset Description.}
The dataset contains pavement inspection records from 2021 to 2024 for $N=750$ road segments. Each record includes the features listed above and the measured PCI value.
Let
$\mathcal{D} = \{(X^{(t)}, Y^{(t)})\}_{t=2021}^{2024}$
where
$X^{(t)} \in \mathbb{R}^{N \times F}, \quad
Y^{(t)} \in \mathbb{R}^{N}.$
Each row corresponds to one pavement segment. The experiments follow a temporal prediction setting where the model is trained and validated using historical observations spanning 2021 to 2023. The final evaluation is performed on the 2024 dataset, using the preceding 2022-23 observations as input to assess the model’s ability to generalize to future pavement condition prediction and simulate real-world decision-making scenarios.

\noindent \textbf{Graph Topology Construction.}
The connectivity and arrangement of a network is known as its topology \citep{ahmadzai2019assessment}. Thus, road transport networks have various specific topologies denoting their structures in terms of edges, vertices, paths, and cycles \citep{rodrigue2020geography}. The road network topology is constructed using adjacency relationships between pavement segments.
If segment $i$ is directly adjacent to segment $j$, an undirected edge is added:
$e_{ij} \in E.$
The graph structure is represented by an edge index matrix
$\mathbf{A} \in \{0,1\}^{N \times N},$
where
\begin{equation} A_{ij} = \begin{cases} 1 & \text{if segments } i \text{ and } j \text{ are adjacent}\\ 0 & \text{otherwise} \end{cases}. \end{equation}
Since adjacency relationships are symmetric, the graph is treated as undirected. The process of graph construction is illustrated in Figure ~\ref{fig2}.

\noindent \textbf{Temporal graph snapshots.}
Each year corresponds to a graph snapshot:
$G^{(t)} = (V, E, X^{(t)}).$
The node set and graph topology remain fixed across time, while node features evolve annually.

\noindent \textbf{Temporal Window Construction.}
To incorporate historical context, a sliding temporal window of length $T_0$ is constructed.
For each training instance,
$\mathbf{X}_i =
\left[
\mathbf{x}_i^{(t-T_0+1)},
\mathbf{x}_i^{(t-T_0+2)},
\ldots,
\mathbf{x}_i^{(t)}
\right]$
and the prediction target becomes
$y_i^{(t+1)}.$
Thus each sample consists of
$\mathbf{X} \in \mathbb{R}^{N \times T_0 \times F}.$

\noindent \textbf{Feature Normalization.}
Feature magnitudes vary substantially (e.g., AADT vs crack percentage). Therefore all features are standardized using
$\tilde{x} = \frac{x - \mu}{\sigma},$
where $\mu$ and $\sigma$ are computed using only training data.
Targets are similarly normalized during training and inverse-transformed during evaluation.

\begin{table}[t]
\centering
\caption{Notations.}
\label{tab:notations}
\begin{tabular}{p{2cm} p{15cm}}
\hline
Symbol & Description \\
\hline
$G$ & Graph representation of the road network $(V,E)$ \\
$V$ & Set of pavement segments (graph nodes) \\
$E$ & Set of adjacency relationships between segments (edges) \\
$N$ & Total number of pavement segments in the network \\
$T$ & Total number of temporal observations (years) \\
$F$ & Number of node features describing pavement properties \\
$v_i$ & $i$-th pavement segment (node) in the graph \\
$\mathbf{x}_i^{(t)}$ & Feature vector of node $i$ at time $t$ \\
$X^{(t)}$ & Node feature matrix at time $t$, $X^{(t)} \in \mathbb{R}^{N \times F}$ \\
$Y^{(t)}$ & Vector of ground truth PCI values at time $t$ \\
$y_i^{(t)}$ & Observed Pavement Condition Index of segment $i$ at time $t$ \\
$\hat{y}_i^{(t)}$ & Predicted PCI value for segment $i$ \\
$\mathbf{A}$ & Adjacency matrix representing graph connectivity \\
$A_{ij}$ & Binary indicator showing whether nodes $i$ and $j$ are adjacent \\
$\mathcal{N}(i)$ & Set of neighboring nodes connected to node $i$ \\
$T_0$ & Temporal window length used for historical observations \\
$\mathbf{X}$ & Temporal feature tensor $ \in \mathbb{R}^{N \times T_0 \times F}$ \\
$W$ & Learnable weight matrix in graph attention layer \\
$\mathbf{a}$ & Attention weight vector in GAT layer \\
$\alpha_{ij}$ & Normalized attention coefficient between nodes $i$ and $j$ \\
$\mathbf{h}_i$ & Spatial embedding of node $i$ after graph attention aggregation \\
$\mathbf{z}_i$ & Residual spatial representation of node $i$ \\
$\mathbf{S}_i$ & Temporal sequence of spatial embeddings for node $i$ \\
$h_t$ & Hidden state of the GRU at time step $t$ \\
$f_\theta$ & Proposed ST-ResGAT model parameterized by $\theta$ \\
$\mathcal{L}$ & Training loss function (Mean Squared Error) \\
$I_f$ & Permutation-based importance score for feature $f$ \\
$\mathbf{M}_f$ & Feature importance mask learned by GNNExplainer \\
$\mathbf{M}_e$ & Edge importance mask learned by GNNExplainer \\
\hline
\end{tabular}
\end{table}

\subsection{ST-ResGAT}

\begin{figure*}[tb]
    \centering
    \includegraphics[width=\textwidth]{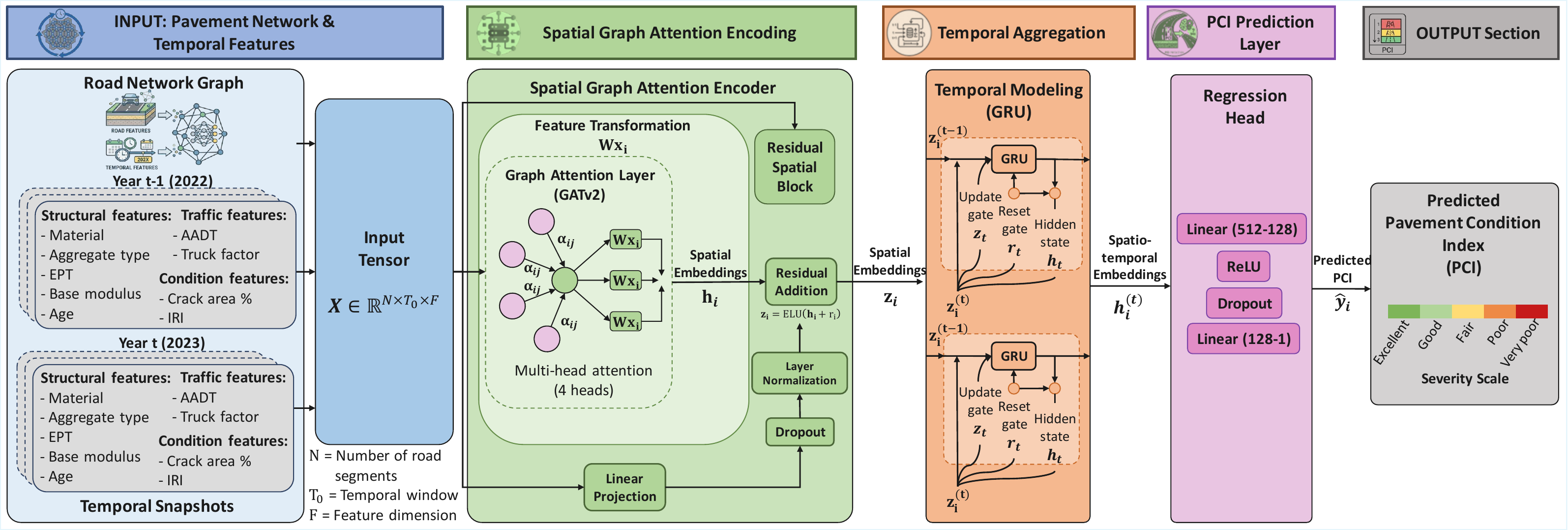}
    \caption{\textbf{Architecture of the proposed Spatio-Temporal Residual Graph Attention Network (ST-ResGAT) for pavement condition prediction}. The model receives temporal node features and the road network graph as input. Spatial dependencies among adjacent pavement segments are learned through multi-head Graph Attention Network (GAT) layer with residual connections to preserve original structural attributes. The resulting spatial embeddings across multiple time steps are aggregated using a Gated Recurrent Unit (GRU) to capture temporal deterioration patterns. The final spatio-temporal representation is passed through a regression head to estimate the Pavement Condition Index (PCI) for each segment, which can subsequently be used for maintenance prioritization.}
    \label{fig1}
\end{figure*}

The proposed model integrates spatial graph attention and temporal sequence modeling to capture both spatial dependencies among adjacent road segments and temporal deterioration patterns. The architecture consists of three main components: (i) a spatial encoder based on Graph Attention Networks that learns interactions among neighboring pavement segments, (ii) a residual spatial representation module that preserves original feature information and stabilizes training, and (iii) a temporal aggregation module implemented using a Gated Recurrent Unit (GRU) to model historical pavement deterioration across multiple years. The architecture of the proposed model is illustrated briefly in Figure ~\ref{fig1}.

\subsubsection{Spatial Graph Attention Encoder}
Road pavement segments are not independent. The condition of a road segment is often influenced by the condition of nearby segments because adjacent pavement sections typically experience similar environmental exposure, traffic loading, and construction characteristics \citep{gao2024considering}. To model this spatial dependency within the road network, a Graph Attention Network (GAT) \citep{vaswani2017attention} layer is used as the spatial encoder.

For each time step $t$, node features are propagated through the graph structure so that each pavement segment can incorporate information from its neighboring segments. Instead of assigning equal influence to all neighbors, the model learns an attention weight that determines how strongly each neighboring segment contributes to the representation of a given segment.
For node $i$, the attention coefficient with neighbor $j$ is computed as

\begin{equation}
e_{ij} =
\text{LeakyReLU}
\left(
\mathbf{a}^{T}
[
W \mathbf{x}_i \, \| \, W \mathbf{x}_j
]
\right)
\end{equation}

where $W$ is a learnable weight matrix that transforms the input features, $\mathbf{a}$ is the attention vector that measures the compatibility between two nodes, and $\|$ denotes feature concatenation. This operation allows the model to compare the feature characteristics of neighboring pavement segments.

The raw attention scores are then normalized across all neighbors of node $i$ using a softmax function:

\begin{equation}
\alpha_{ij} =
\frac{\exp(e_{ij})}
{\sum_{k \in \mathcal{N}(i)} \exp(e_{ik})}.
\end{equation}

These normalized coefficients determine how much influence each neighbor has when updating the representation of node $i$. The updated node representation is obtained by aggregating the transformed features of its neighbors weighted by the learned attention coefficients:

\begin{equation}
\mathbf{h}_i =
\sigma
\left(
\sum_{j \in \mathcal{N}(i)}
\alpha_{ij} W \mathbf{x}_j
\right).
\end{equation}

To improve the expressive capacity of the spatial encoder, multi-head attention is used. Multiple attention mechanisms operate in parallel, each learning a different interaction pattern among neighboring segments. The outputs of these heads are concatenated:

\begin{equation}
\mathbf{h}_i =
\Vert_{k=1}^{K}
\sigma
\left(
\sum_{j \in \mathcal{N}(i)}
\alpha_{ij}^{(k)} W^{(k)} \mathbf{x}_j
\right).
\end{equation}

This multi-head mechanism allows the model to capture different spatial relationships, such as shared traffic loading patterns or similar structural characteristics among adjacent pavement segments.

\subsubsection{Residual Spatial Representation}
While graph attention layers effectively capture spatial interactions, deeper graph transformations may sometimes distort the original node features \citep{wu2023demystifying}. In pavement condition modeling, the original structural and traffic attributes remain important predictors and should not be lost during feature propagation. To preserve this information and improve training stability, a residual connection is incorporated into the spatial encoder.

The original node features are first projected into the same embedding dimension using a linear transformation:
$\mathbf{r}_i = W_r \mathbf{x}_i$
This residual representation is then added to the attention-based embedding:
\begin{equation}
\mathbf{z}_i = \text{ELU}(\mathbf{h}_i + \mathbf{r}_i).
\end{equation}

The residual connection ensures that the model retains access to the raw structural and traffic attributes while also incorporating information aggregated from neighboring segments. After this step, layer normalization and dropout are applied to stabilize training and reduce overfitting.

\subsubsection{Temporal Feature Aggregation}
Pavement deterioration is inherently a temporal process. The structural condition of a road segment gradually changes over time due to traffic loading, environmental exposure, and material aging. Therefore, it is important for the model to capture how pavement features evolve across multiple years rather than relying on a single snapshot.

For each node, the spatial embeddings obtained from the previous steps are collected across the temporal window:
$\mathbf{S}_i =
[
\mathbf{z}_i^{(t-T_0+1)},
\ldots,
\mathbf{z}_i^{(t)}
].$
This sequence describes how the structural and spatial characteristics of a pavement segment evolve over time. The sequence is then processed using a Gated Recurrent Unit (GRU), which is a recurrent neural network architecture designed to capture temporal dependencies while avoiding vanishing gradient issues.

The GRU maintains a hidden state that summarizes historical information. Its update process is governed by two gates that control how information flows through time.

The update gate is computed as
\begin{equation}
\mathbf{z}_t = \sigma(W_z x_t + U_z h_{t-1})
\end{equation}

which determines how much of the previous hidden state should be retained.

The reset gate is computed as

\begin{equation}
\mathbf{r}_t = \sigma(W_r x_t + U_r h_{t-1})
\end{equation}

which determines how strongly past information influences the candidate hidden state.

The candidate hidden representation is then calculated as

\begin{equation}
\tilde{h}_t =
\tanh(W_h x_t + U_h (r_t \odot h_{t-1}))
\end{equation}

and the final hidden state is updated as

\begin{equation}
h_t = (1 - z_t) \odot h_{t-1} + z_t \odot \tilde{h}_t.
\end{equation}

Through these gating mechanisms, the GRU learns how pavement characteristics evolve across years and how past conditions influence future deterioration.
The final hidden state
$h_i^{(t)}$
acts as a compact representation summarizing both spatial interactions and historical pavement evolution for each road segment.

\subsubsection{Prediction Layer}
The final step of the model is to estimate the Pavement Condition Index for the next time period. The temporal representation obtained from the GRU is passed through a feed-forward regression network that maps the learned embedding to a scalar PCI prediction.

The predicted PCI value for node $i$ is computed as

\begin{equation}
\hat{y}_i =
W_2
\phi
(
W_1 h_i^{(t)}
)
\end{equation}

where $\phi(\cdot)$ denotes the ReLU activation function.

This regression layer learns a mapping between the combined spatial–temporal representation and the expected pavement condition value. The resulting prediction reflects both the current structural characteristics of the segment and the historical deterioration patterns observed across the network.

\subsubsection{Learning Objective}
The model is trained using Mean Squared Error loss computed over training nodes:
\begin{equation}
\mathcal{L} =
\frac{1}{|V_{train}|}
\sum_{i \in V_{train}}
(\hat{y}_i - y_i)^2.
\end{equation}

\subsubsection{Training Procedure}
Training proceeds by first constructing temporal graph windows from historical observations. For each window, the model receives node feature sequences together with the corresponding graph topology. Spatial representations of pavement segments are computed using graph attention layers, which capture interactions among adjacent road segments. These spatial embeddings across the temporal window are then processed by the GRU module to model deterioration dynamics over time. The resulting temporal representation is passed through the regression head to predict the PCI value for the subsequent year. Model parameters are updated through gradient-based optimization using the mean squared error loss between the predicted and observed PCI values.

\subsubsection{Prediction-based Maintenance Prioritization}

To ensure practical applicability, the continuous PCI predictions were translated into discrete actionable categories using the globally recognized ASTM D6433 \citep{ASTM-D6433-23} standard as illustrated in Table ~\ref{tab1}.

\begin{table}[htbp]
    \centering
    \caption{Pavement condition index (PCI) severity and recommendations according to \cite{ASTM-D6433-23}}
    \label{tab2}
        \begin{tabular}{ccll}
            \toprule
            \textbf{PCI} & \textbf{Severity Rank} & \textbf{Recommended Action} & \textbf{Physical Implication} \\
            \midrule
            86--100 & 1 (Very low) & Routine monitoring  & Structurally sound, No immediate action \\
            71--85  & 2 (Low)      & Preventive          & Noticeable wear, Crack sealing required \\
            56--70  & 3 (Moderate) & Corrective          & Significant distress, Patching required \\
            41--55  & 4 (High)     & Major overlay       & Severe distress, Potential base failure \\
            0--40   & 5 (Critical) & Full reconstruction & Structural failure, High safety hazard \\
            \bottomrule
        \end{tabular}%
\end{table}

Let $\hat{y}_i$ denote the predicted PCI.
Severity categories are assigned using predefined thresholds:
\begin{equation}
S_i =
\begin{cases}
\text{Excellent} & 85 \le \hat{y}_i \le 100 \\
\text{Good} & 70 \le \hat{y}_i < 85 \\
\text{Fair} & 55 \le \hat{y}_i < 70 \\
\text{Poor} & 40 \le \hat{y}_i < 55 \\
\text{Very Poor} & \hat{y}_i < 40
\end{cases}.
\end{equation}

The predicted severity rankings are compared against rankings derived from actual PCI values. Although exact PCI values may differ slightly, the predicted categories largely align with the actual prioritization levels. This indicates that the ST-ResGAT model captures the deterioration patterns sufficiently well to support practical maintenance planning.

\subsection{Explainable AI}
Although the proposed ST-ResGAT model achieves high predictive performance, interpretability is necessary to understand which pavement characteristics influence the predicted condition scores. To address this, we employ an explainable AI framework combining \textbf{GNNExplainer} \citep{ying2019gnnexplainer}for local explanations and a \textbf{permutation-based feature importance} method for global interpretability.

\subsubsection{Local Explanation using GNNExplainer}
Graph neural networks operate on complex graph structures, making it difficult to interpret the contribution of individual features and edges \citep{lu2025modeling}. To analyze the model's decision process for a specific pavement segment, we utilize GNNExplainer.

Given a trained model $f_\theta$ and a node $v_i$, GNNExplainer seeks a minimal subgraph $G_S$ and feature subset $F_S$ that maximizes the mutual information between the model prediction and the explanation:
$\max_{G_S, F_S} I(Y_i ; G_S, F_S)$
where
$Y_i = f_\theta(G, X)_i$
represents the predicted PCI for node $i$.
The explanation process learns two masks: \textit{(i) Feature mask} $\mathbf{M}_f$ identifying important node attributes
\textit{(ii) Edge mask} $\mathbf{M}_e$ identifying influential graph connections.
These masks are optimized through gradient-based learning:

\begin{equation}
\min_{\mathbf{M}_f,\mathbf{M}_e}
\mathcal{L}_{pred}
+ \lambda_1 ||\mathbf{M}_f||_1
+ \lambda_2 H(\mathbf{M}_f)
+ \lambda_3 ||\mathbf{M}_e||_1
\end{equation}
where $H(\cdot)$ denotes entropy regularization encouraging sparse explanations.

\subsubsection{Adapting GNNExplainer to Spatio-Temporal Graph Inputs}
The ST-ResGAT model accepts temporal feature sequences
$X \in \mathbb{R}^{N \times T_0 \times F}$
where $T_0$ represents the historical time window. However, the GNNExplainer framework assumes static node features of dimension
$X \in \mathbb{R}^{N \times F}.$

To bridge this mismatch, a wrapper model is introduced that converts static node features into temporal sequences.
Let
$\mathbf{x}_i \in \mathbb{R}^{F}$
denote the perturbed node features provided by GNNExplainer. The wrapper reconstructs the temporal sequence by repeating the feature vector across the temporal dimension:
$\tilde{X}_i =
[
\mathbf{x}_i,
\mathbf{x}_i,
\ldots,
\mathbf{x}_i
]$
for $T_0$ time steps. The resulting tensor
$\tilde{X} \in \mathbb{R}^{N \times T_0 \times F}$
is then forwarded through the ST-ResGAT model.
This strategy allows GNNExplainer to evaluate feature importance for the spatial component of the model while preserving compatibility with the temporal architecture.

\subsubsection{Node-level Feature Importance}
For a selected pavement segment $v_i$, GNNExplainer produces a feature importance vector
$\mathbf{m}_i \in \mathbb{R}^{F}$
indicating the relative contribution of each attribute to the predicted PCI.
To facilitate visualization and comparison, the importance values are normalized:

\begin{equation}
\tilde{m}_{i,f} =
\frac{m_{i,f}}{\sum_{k=1}^{F} m_{i,k}}
\end{equation}

where $\tilde{m}_{i,f}$ represents the normalized importance of feature $f$.

These normalized scores are visualized using bar plots to illustrate which structural, traffic, or condition variables most strongly influence predictions for a given road segment.

\subsubsection{Global Feature Importance}
While GNNExplainer provides local explanations, global feature importance is assessed using a permutation-based approach \citep{hassija2024interpreting}.
Let $\hat{Y}$ denote model predictions and $Y$ denote ground truth PCI values. The baseline prediction error is computed using Mean Squared Error (MSE):

\begin{equation}
\text{MSE}_{base} =
\frac{1}{N}
\sum_{i=1}^{N}
(y_i - \hat{y}_i)^2
\end{equation}

To measure the importance of feature $f$, the feature values are randomly permuted across nodes:
$X_{:, :, f}^{perm}$
which disrupts the relationship between that feature and the target variable.

The model predictions are recomputed using the perturbed features, producing a new error value:
$\text{MSE}_f^{perm}$.
The feature importance score is defined as
$I_f =
\text{MSE}_f^{perm} - \text{MSE}_{base}.$
A larger increase in error indicates greater reliance on that feature.

To obtain relative contributions, the scores are normalized:
\begin{equation}
\tilde{I}_f =
\frac{I_f}{\sum_{k=1}^{F} I_k}.
\end{equation}

This method quantifies the overall influence of each structural and traffic feature on pavement health prediction.

\section{Data and Experimental Setup}\label{sec5}

\subsection{Data Collection} \label{sec:5.1}

To evaluate the efficacy of the proposed ST-ResGAT framework, we utilize a real-world, high-resolution dataset encompassing 750 physical road segments within the Sylhet region of Bangladesh. The data was meticulously curated from the official records of the Roads and Highway Department (RHD) \citep{rhd}. For the data collection, a roughometer III mounted on a vehicle is used by the \citep{rhd} of Bangladesh. The Roads and Highways Department (RHD) calculates the Pavement Condition Index (PCI) following the methodology outlined in ASTM D6433 \citep{ASTM-D6433-23}, which provides a standardized numerical rating (0–100) for roadway condition. The process begins with a visual survey of defined sample units to identify the type, severity, and quantity of surface distresses. These observations are converted into Deduct Values using density-based weighting factors that represent the impact of each defect. To account for multiple distresses without over-penalizing the score, a Corrected Deduct Value (CDV) is iteratively calculated, and the final PCI is determined by subtracting the maximum CDV from 100 \citep{zafar2019condition}. This data is central to the RHD Road Maintenance Management System (RMMS). The rest of the input features were collected from the \cite{rhd} database. Sylhet was strategically selected as the study area due to its unique socio-environmental profile. It is one of the most flood-vulnerable regions in the country, frequently subjected to flash floods that accelerate pavement deterioration, while simultaneously serving as a critical national tourist hub. Maintaining infrastructure health in this region is therefore paramount for both disaster resilience (SDG 11/13) and the local tourism-driven economy (SDG 8).

\subsection{Data Statistics} \label{sec:5.2}
The compiled dataset represents a longitudinal pavement monitoring record spanning the years 2021 to 2024. The network consists of 750 road segments represented as graph nodes, with 3000 temporal observations in total. Each node corresponds to a spatially distinct pavement segment, while edges represent adjacency relationships between neighboring segments. In total, the constructed road graph contains 1498 bidirectional adjacency connections, enabling the model to capture spatial dependencies across the transportation network.

Each node observation is characterized by twelve explanatory variables describing the structural, traffic, environmental, and material conditions influencing pavement deterioration. Structural attributes include existing pavement thickness (EPT), base modulus, and pavement age in years. Traffic-related loading conditions are represented through Annual Average Daily Traffic (AADT) and the truck factor, both of which influence fatigue accumulation and structural wear. Material characteristics are captured through categorical indicators for pavement material type and aggregate type. Environmental exposure is represented by flood risk level and proximity to quarry sources, while surface distress indicators include the International Roughness Index (IRI) and crack area percentage. The Pavement Condition Index (PCI) serves as the primary prediction target and represents the overall structural health of each pavement segment. Table~\ref{tab:dataset_stats} summarizes the descriptive statistics of all variables used in the modeling process.

\begin{table}[htbp]
\centering
\caption{Descriptive statistics of the dataset.}
\label{tab:dataset_stats}
\begin{tabular}{lcccccc}
\hline
Feature & Unique & Min & Max & Mean & Std \\
\hline
Material & 2 & 0 & 1 & 0.74 & 0.44 \\
Agg\_Type & 2 & 0 & 1 & 0.87 & 0.34 \\
Flood\_Risk & 3 & 0 & 2 & 0.93 & 0.86 \\
Proximity\_Quarry & 2 & 0 & 1 & 0.26 & 0.44 \\
Age\_Yrs & 10 & 2 & 11 & 6.56 & 2.28 \\
Traffic\_AADT & 2505 & 5004 & 19977 & 9481.24 & 4039.14 \\
Truck\_Factor & 817 & 1.00 & 11.99 & 5.77 & 3.09 \\
EPT\_mm & 538 & 120.60 & 320.00 & 253.80 & 58.69 \\
Base\_Modulus & 657 & 150.80 & 499.60 & 349.56 & 103.09 \\
Crack\_Area\_Pct & 1069 & 0.00 & 21.92 & 4.32 & 3.95 \\
IRI & 349 & 1.00 & 5.66 & 2.65 & 0.69 \\
PCI & 1950 & 36.65 & 97.05 & 80.99 & 9.13 \\
\hline
\end{tabular}
\end{table}

The spatial structure of the dataset is represented through a directed adjacency graph constructed from road connectivity information. Each edge represents a direct adjacency relationship between two neighboring road segments, allowing the graph neural network to propagate structural and environmental signals across the network topology. Table~\ref{tab:graph_summary} provides a summary of the graph characteristics.

\begin{table}[htbp]
\centering
\caption{Graph structure of the road network dataset.}
\label{tab:graph_summary}
\begin{tabular}{lc}
\hline
Property & Value \\
\hline
Road segments (nodes) & 750 \\
Temporal observations & 3000 \\
Edges (adjacent connections) & 1498 \\
Time span & 2021--2024 \\
Connection type & Adjacent \\
\hline
\end{tabular}
\end{table}

\subsection{Temporal Data Partitioning}
To ensure realistic forecasting evaluation and prevent information leakage, the dataset was partitioned chronologically rather than randomly. This strategy mimics real-world infrastructure forecasting scenarios in which models are trained on historical records and deployed to predict future pavement conditions.
The first two years of observations (2021–2022), representing approximately 50\% of the temporal timeline, were used as the training set. During this phase, the model optimized the parameters of the spatial attention layers and the temporal aggregation modules to learn baseline deterioration patterns across the network.
The subsequent year (2023), corresponding to 25\% of the temporal records, served as the validation set. This subset was used for architectural ablation experiments and hyperparameter tuning, including the calibration of attention heads, hidden feature dimensions, and temporal aggregation parameters.
Finally, the most recent observations from 2024 were reserved as an unseen test set. This partition provides a realistic benchmark for evaluating the model’s predictive performance under forward-looking conditions and reflects the practical scenario in which transportation agencies deploy predictive models to forecast upcoming pavement states using previously observed infrastructure data.

\subsection{Data Pre-processing}

In this work, we adopt a temporal graph learning approach where nodes (road segments) and edges (physical adjacency) remain static, while node features evolve over time to reflect dynamic infrastructure and environmental attributes. The goal is to predict the future structural integrity of the pavement, specifically the Pavement Condition Index (PCI), using these multi-physics temporal signals. We preprocess the raw data through a multi-stage pipeline, applying feature scaling to normalize heterogeneous variables such as high-magnitude traffic volumes (AADT) and fractional climatic stressors (Flood Risk), ensuring numerical stability during gradient descent. Categorical attributes, including pavement and aggregate types, were encoded to ensure consistency across all time steps. Finally, rather than relying on out-of-the-box library wrappers, the dataset was engineered into custom spatio-temporal data structures. Using a sliding window approach ($T_0=2$), the sequences were reshaped into 3D tensors ($Nodes$ × $T_{0}$ × $Features$) and paired with their static spatial adjacency matrices, enabling highly customized and efficient training of our ST-ResGAT architecture.

\subsection{Baseline Models}

We compare the proposed ST-ResGAT framework with various traditional machine learning, deep learning, and graph-based architectural models to rigorously evaluate its predictive capabilities. Among these, we utilize several robust ensemble learning algorithms widely applied in infrastructure deterioration modeling, including Random Forest (RF) \citep{breiman2001random}, Extreme Gradient Boosting (XGBoost) \citep{chen2016xgboost}, Categorical Boosting (CatBoost) \citep{Prokhorenkova2018Catboost}, and Adaptive Boosting (AdaBoost) \citep{freund1996experiments}. For deep learning-based (DL) baselines, we consider the Multilayer Perceptron (MLP) \citep{kruse2022multi} to evaluate the predictive performance of a standard, fully connected feed-forward neural network operating without spatial or temporal inductive biases. Furthermore, to isolate and validate the specific contributions of our framework's core mathematical components, we compare ST-ResGAT against its own ablated variants. These graph-based baselines include the Vanilla Graph Attention Network (GAT) \citep{velivckovic2018graph} as a spatial-only baseline, ST-GAT to test the model without residual connections, and ResGAT to evaluate performance without the GRU-based temporal memory. All baseline models were rigorously evaluated using their optimally tuned hyperparameters. To ensure a completely fair comparative analysis, the input feature structure, data partitioning, and temporal sequence lookback window ($T_0=2$) used across all sequential and non-sequential baselines are identical to that of ST-ResGAT, with the sole distinction being how each architecture internally routes and processes the multi-physics data to predict the Pavement Condition Index (PCI).

\subsection{Evaluation Metrics}

To assess the performance of the pavement condition prediction model, we employ multiple standard regression evaluation metrics that collectively offer a comprehensive view of prediction accuracy. Mean Squared Error ($ MSE = \frac{1}{n}\sum_{i=1}^{n}(y_i - \hat{y}_i)^2$) computes the average of the squared differences between the predicted and observed Pavement Condition Index values, inherently applying a harsher mathematical penalty to larger predictive discrepancies \citep{wang2009mean}. To contextualize this error magnitude within the original scale of the target variable, we calculate the Root Mean Squared Error ($RMSE = \sqrt{MSE}$), which significantly improves interpretability for practical infrastructure condition assessment \citep{chai2014root}.  Additionally, the Mean Absolute Error ($ MAE = \frac{1}{n}\sum_{i=1}^{n}|y_i - \hat{y}_i|$) evaluates the arithmetic mean of absolute forecasting errors, supplying a direct measure of average deviation that remains robust against statistical outliers \citep{botchkarev2019new}. Finally, the Coefficient of Determination ($R^2 = 1 - \frac{\sum_{i=1}^{n}(y_i - \hat{y}_i)^2}{\sum_{i=1}^{n}(y_i - \bar{y}_i)^2}$) quantifies the ratio of variance in pavement deterioration that the model successfully explains, where an outcome approaching unity signifies exceptional predictive alignment with the ground truth \citep{piepho2019coefficient}. By synthesizing these indicators, we establish a comprehensive foundation for comparing both the raw accuracy and the structural reliability of our model against competing methodologies.

\subsection{Immplementation Details}

\noindent \textbf{Software Framework.}
The proposed ST-ResGAT model is implemented using the PyTorch deep learning framework together with the PyTorch Geometric library \citep{fey2019fast} for graph neural network operations. Graph attention layers are implemented using the \texttt{GATv2Conv} operator \citep{shi2025two}. Data preprocessing and evaluation are conducted using NumPy, Pandas, and Scikit-learn \citep{harris2020array}. A deterministic training setup is used where random seeds are fixed across Python, NumPy, and PyTorch to ensure reproducibility. Model training utilizes GPU acceleration when available, otherwise the computation is executed on CPU.

\noindent \textbf{Graph Construction.}
The road network graph contains 750 pavement segments, where each segment is represented as a node with 11 input features. The node ordering is kept consistent across all yearly snapshots by constructing a canonical sorted list of segment identifiers. The adjacency relationships between pavement segments define the graph edges, and an undirected representation is created by inserting both directions for each edge pair.

\noindent \textbf{Temporal Sequence Preparation.}
Temporal sequences are constructed using a sliding window with a temporal history length of $T_0 = 2$. Each input sample therefore consists of node feature sequences from two consecutive years. For example, features from 2022 and 2023 are used to predict pavement condition for 2024. This process results in temporal training pairs where the model receives a feature tensor of shape $N \times T_0 \times F$ together with the shared graph topology.

\noindent \textbf{Feature Normalization.}
All node features are standardized using a standard scaling procedure fitted on the training data. The same transformation parameters are applied across all temporal steps. Target PCI values are also standardized during training and later transformed back to the original scale for evaluation.

\noindent \textbf{Spatial Encoder Configuration.}
The spatial encoder uses a graph attention layer with a hidden dimension of 128 per attention head and 4 attention heads, resulting in a spatial embedding dimension of 256. The ELU activation function is applied after the attention operation, followed by layer normalization and dropout. A residual linear projection is used to align the original feature space with the spatial embedding dimension before the residual addition. The dropout probability used in the spatial module is 0.

\noindent \textbf{Temporal Aggregation Module.}
Temporal aggregation is performed using a single-layer Gated Recurrent Unit (GRU). The GRU receives spatial embeddings of dimension 256 for each time step and produces a hidden representation of dimension 256 for each node. The final hidden state of the GRU serves as the temporal representation used for PCI prediction.

\noindent \textbf{Prediction Head.}
The regression head consists of a fully connected layer mapping the GRU hidden representation from 256 to 128 dimensions, followed by a ReLU activation and a dropout layer with probability 0. A final linear layer maps the 128-dimensional representation to a single scalar output corresponding to the predicted PCI value.

\noindent \textbf{Optimization Settings.}
Model optimization is performed using the Adam optimizer with an initial learning rate of $10^{-3}$ and $10^{-4}$ weight decay. Training is conducted for a maximum of 200 epochs. A ReduceLROnPlateau learning rate scheduler is used to reduce the learning rate when validation performance stagnates, with a reduction factor of 0.5 and a patience of 8 epochs.

\noindent \textbf{Training Strategy.}
The loss is computed only on the nodes belonging to the training subset while validation performance is monitored using the validation nodes. Early stopping is applied with a patience of 25 epochs to prevent overfitting. The model parameters corresponding to the best validation performance are saved and used for final evaluation.

\noindent \textbf{Testing Procedure.}
During testing, the trained model receives the temporal sequence constructed from the two preceding years and predicts PCI values for all nodes in the graph. The predictions are then transformed back to the original PCI scale and compared against the ground truth values for evaluation.

\section{Experimental Results}\label{sec5}
This section presents a comprehensive analysis of the prediction results obtained from various baseline and the proposed model. In this study, the comparative performance across standard evaluation metrics, interpretation of the “black-box” model, and the key aspects of the findings support the SDGs are discussed. A detailed ablation study for different components and model hyperparameters is also included in this study.

\subsection{Quantitative Results}

The results in Table ~\ref{tab3}, reveal the comparative predictive capabilities of the evaluated models on the unseen test set. The proposed ST-ResGAT framework establishes a new state-of-the-art for this forecasting task, achieving a substantially lower Mean Squared Error (MSE) of 7.4096 and capturing an exceptional 93.22\% of the variance in pavement deterioration ($R^2$ = 0.9322). Crucially, a comparative analysis against the baseline methods reveals the fundamental limitations of traditional, non-graph paradigms. Both the standard neural network (MLP) and the strongest ensemble machine learning model (CatBoost) appear to hit a strict performance ceiling, plateauing around an $R^2$ of 0.88 with MSEs exceeding 12.1. By explicitly abandoning the assumption that pavement segments deteriorate as isolated, independent entities, the ST-ResGAT yields a striking ~38.8\% reduction in MSE compared to these best-performing baselines. Furthermore, the particularly poor generalization of conventional tree-based models like Random Forest and AdaBoost underscores their inability to extrapolate complex temporal degradation trajectories. Ultimately, these results empirically validate that integrating explicit spatio-temporal network topology is not merely an incremental enhancement, but a critical prerequisite for robust, high-fidelity infrastructure forecasting.

\begin{table}[htbp]
    \centering
    \caption{Performance metrics comparison on test set}
    \label{tab3}
        \begin{tabular}{llcccc}
            \toprule
            \textbf{Type} & \textbf{Model} & \textbf{MSE} ($\downarrow$) & \textbf{RMSE} ($\downarrow$) & \textbf{MAE} ($\downarrow$) & \textbf{R\textsuperscript{2}}($\uparrow$) \\
            \midrule
            \multirow{4}{*}{ML} 
             & RandomForest & 19.9238 & 4.4636 & 3.4266 & 0.8177 \\
             & XGBoost      & 15.0562 & 3.8802 & 2.9291 & 0.8622 \\
             & AdaBoost     & 16.5347 & 4.0663 & 3.1908 & 0.8487 \\
             & CatBoost     & 12.1068 & 3.4795 & 2.6313 & 0.8892 \\
            \midrule
            Neural Network 
             & MLP          & 12.2046 & 3.4935 & 2.6632 & 0.8883 \\
            \midrule
            Proposed model 
             & \textbf{ST-ResGAT} & \textbf{7.4096} & \textbf{2.7221} & \textbf{2.0886} & \textbf{0.9322} \\
            \bottomrule
        \end{tabular}%
\end{table}

\begin{figure*}[tb]
    \centering
    \includegraphics[width=\textwidth]{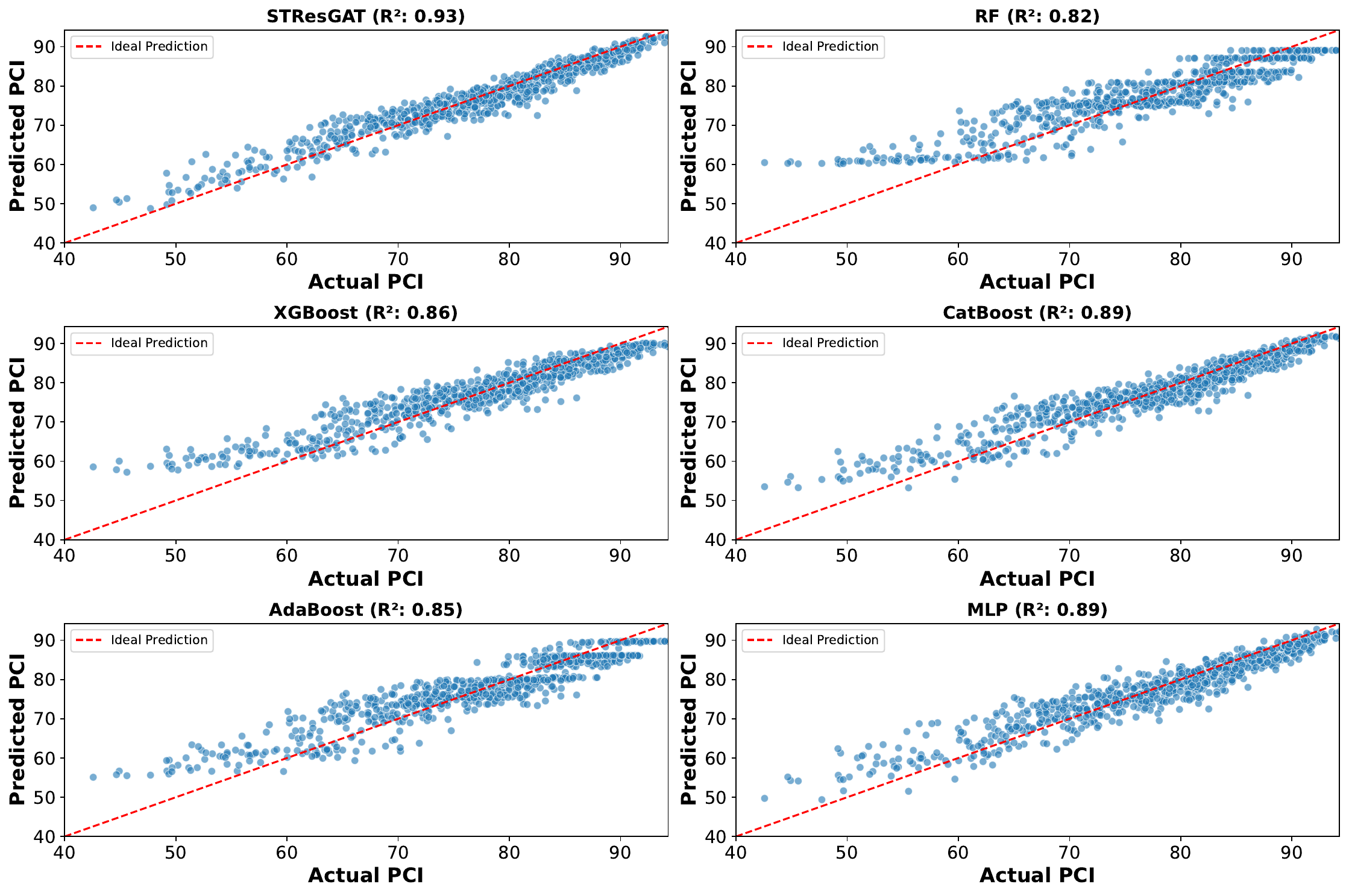}
    \caption{Actual vs. Predicted PCI scatter plot for all comparative models. The ResGAT model 
shows the tightest clustering along the diagonal identity line.}
    \label{fig3}
\end{figure*}

Figure ~\ref{fig3} illustrates the scatter plots of predicted versus actual PCI values for the evaluated models, with the red dashed line denoting the ideal 1:1 relationship. ST-ResGAT demonstrates the closest adherence to the ideal line, indicating superior predictive accuracy and minimal dispersion. CatBoost and MLP exhibit comparable performance with tightly clustered predictions, while XGBoost and AdaBoost show slightly increased variance. RF presents comparatively larger deviations from the ideal trend.

\subsection{Dual-Diagnostic Model Appraisal}

\begin{figure*}[tb]
    \centering
    
    \begin{minipage}{0.52\textwidth}
        \centering
        \includegraphics[width=\linewidth]{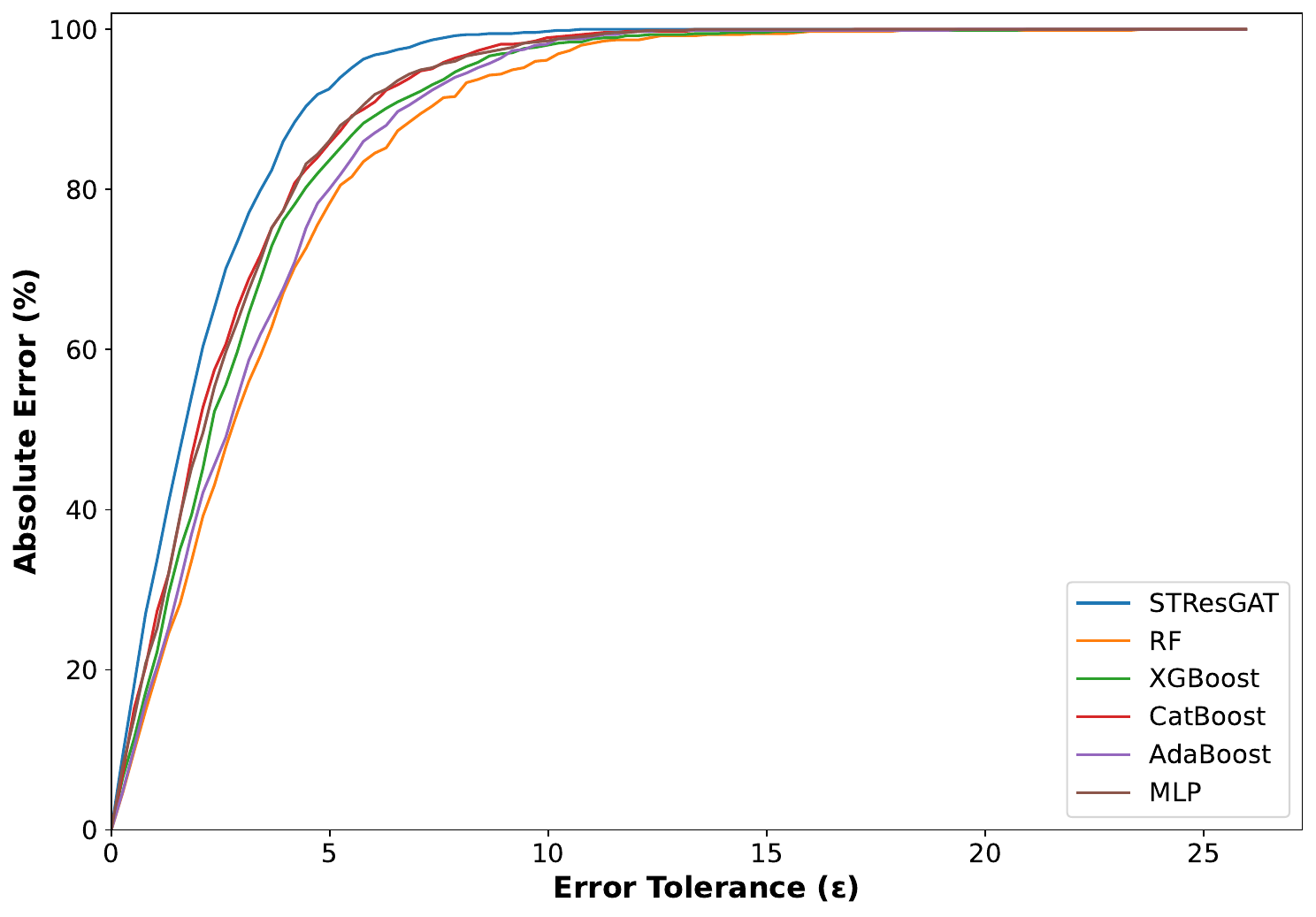}
        \centerline{(a)}
        \label{fig4a}
    \end{minipage}
    \hfill
    \begin{minipage}{0.45\textwidth}
        \centering
        \includegraphics[width=\linewidth]{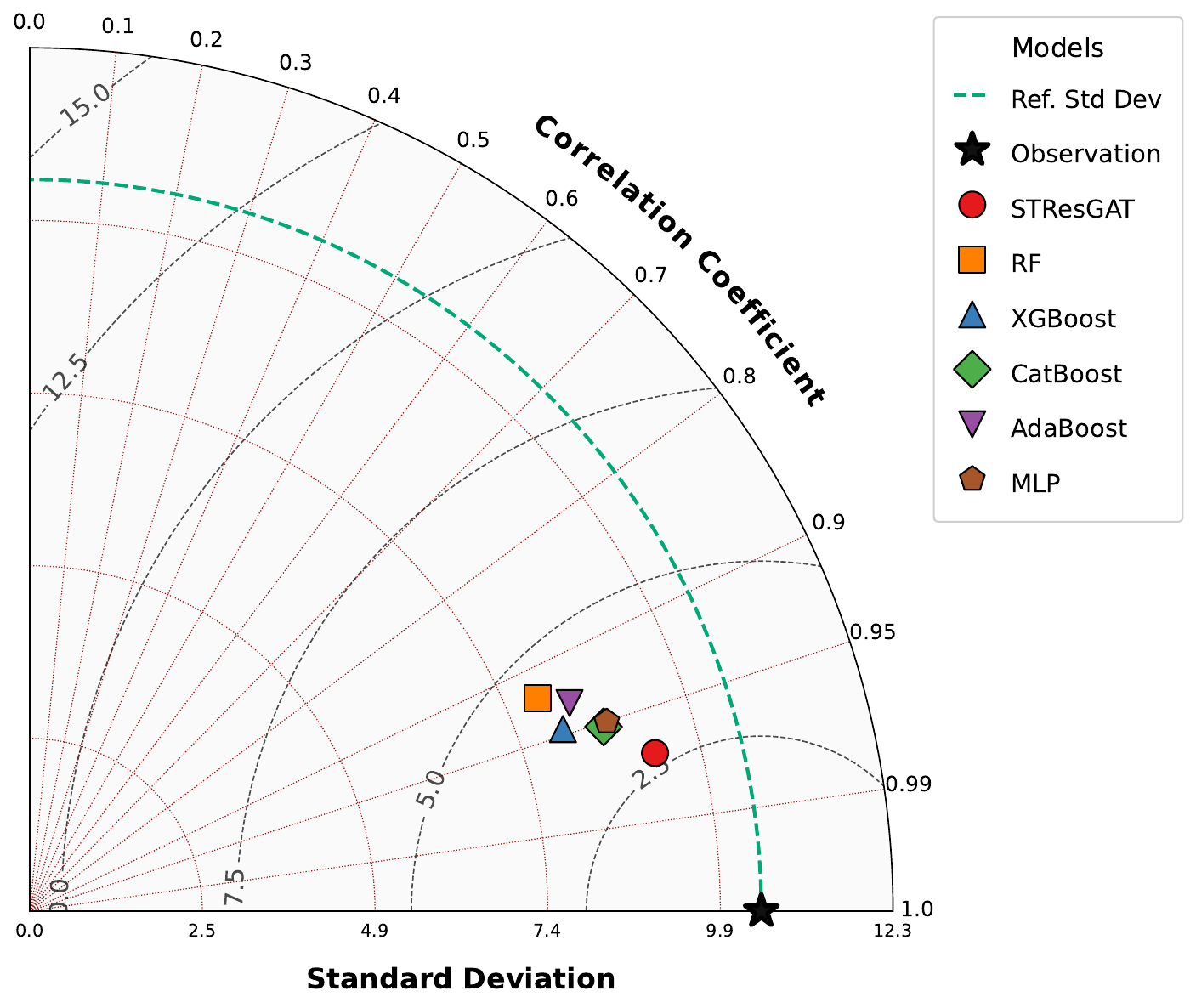}
        \centerline{(b)}
        \label{fig4b}
    \end{minipage}
    
    
    \caption{(a) REC curve, further validating the high error tolerance and robustness of the ST-ResGAT predictions, (b) Taylor Diagram illustrating the standard deviation, root mean square error (RMSE), and correlation coefficient of the models relative to the ground truth observation point.}
    \label{fig4}
\end{figure*}

A synergistic graphical assessment based on Regression Error Curve (REC) curves and Taylor diagrams was conducted to ensure robust and comparative evaluation of model behavior. Figure ~\ref{fig4}a presents the REC curves for all models, illustrating the cumulative percentage of predictions within increasing error tolerances ($\epsilon$). ST-ResGAT consistently dominates across the full error spectrum, achieving higher accuracy at lower $\epsilon$ thresholds, which indicates superior precision and robustness. CatBoost and MLP demonstrate closely competitive performance, followed by XGBoost and AdaBoost with moderate deviations. RF exhibits comparatively slower error convergence, reflecting reduced predictive reliability at stricter tolerances. As $\epsilon$ increases, all models asymptotically approach complete coverage; however, the earlier saturation of ST-ResGAT confirms its overall advantage in predictive consistency.

While REC curves provide complementary assessments of predictive performance, it primarily emphasizes point-wise agreement and error tolerance behavior. It does not, however, simultaneously synthesize correlation structure, variance representation, and centered RMSE within a unified geometric framework. To address this limitation and enable a more holistic evaluation of model skill, a Taylor diagram is further employed. Figure ~\ref{fig4}b illustrates summarizing model performance against observations. ST-ResGAT is positioned closest to the reference point, indicating the highest correlation, minimal centered RMSE, and a standard deviation most consistent with the observed variability. CatBoost and MLP exhibit comparable skill with slightly larger deviations in variance and error magnitude. XGBoost and AdaBoost demonstrate moderate dispersion from the reference, whereas RF shows comparatively lower correlation and greater deviation from the observed standard deviation. Overall, the Taylor diagram corroborates previous findings while additionally confirming that ST-ResGAT most effectively captures both the magnitude and variability structure of the observed PCI data.

\subsection{Proactive Maintenance Profiling and Prioritization}

The central objective of this study is to operationalize predictive modeling for proactive pavement maintenance planning. To this end, the 2024 Pavement Condition Index (PCI) was predicted at the segment level using the ST-ResGAT model and systematically translated into actionable maintenance decisions through an ASTM D6433 \citep{ASTM-D6433-23} -compliant severity ranking framework. This step represents the core practical contribution of the work, moving beyond performance prediction toward structured intervention planning.

\subsubsection{Predictive-to-Decision Translation Stage}

\begin{figure*}[tb]
    \centering
    \includegraphics[width=\textwidth]{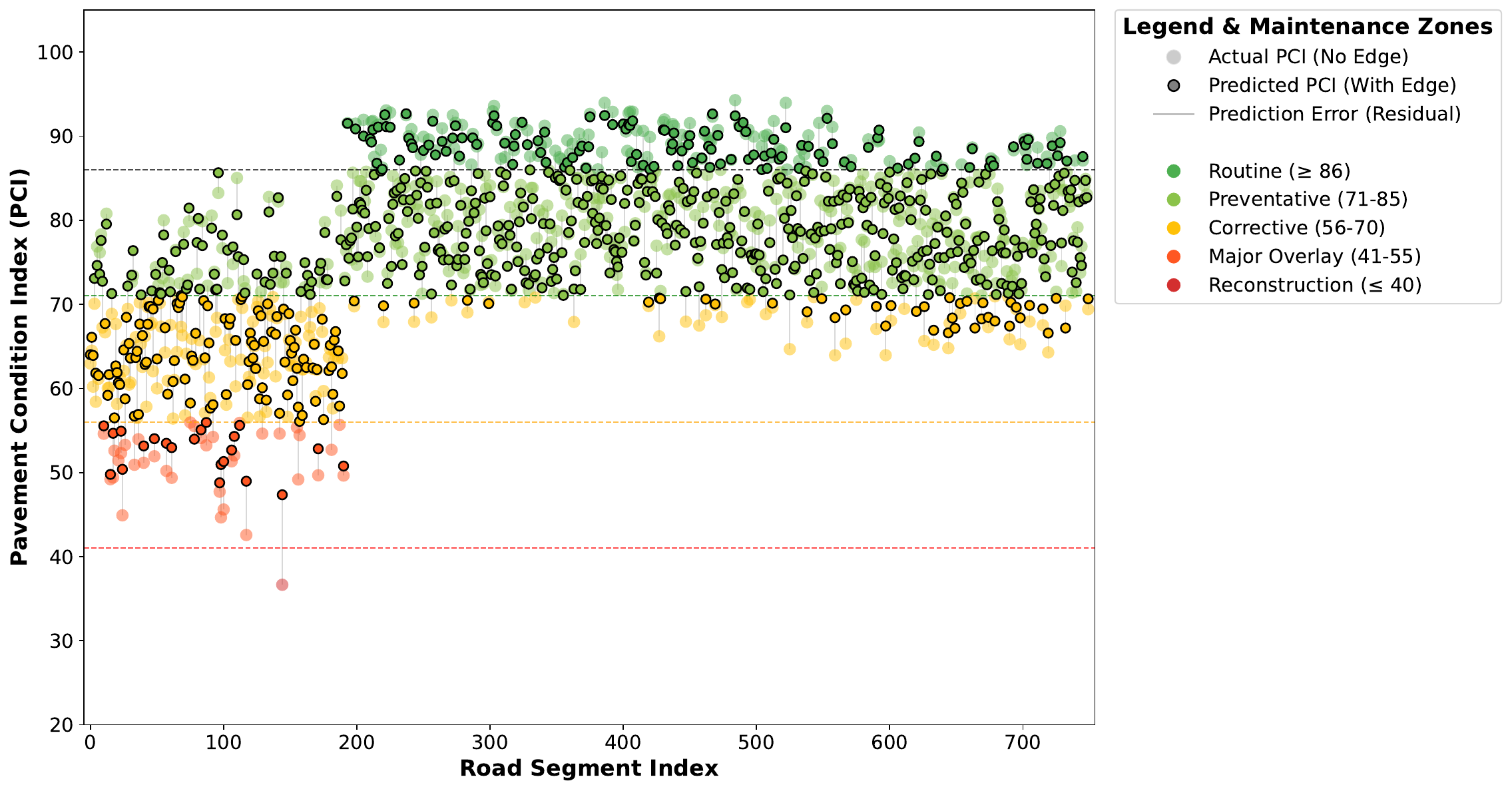}
    \caption{Longitudinal Profile of the maintenance interventions for all 750 segements (2024), illustrating the ST-ResGAT predicted spatial degradation trends against actual conditions, overlaid with ASTM D6433 action thresholds.}
    \label{fig5}
\end{figure*}

For each road segment, the predicted 2024 PCI was first aligned with the corresponding observed PCI to ensure segment-level consistency and to quantify residual deviations. The predicted values were then converted into standardized maintenance categories based on ASTM D6433 PCI thresholds as mentioned in Table ~\ref{tab2}. This categorical transformation enables direct interpretation of continuous PCI predictions within an engineering decision context. The resulting maintenance profile illustrated in Figure ~\ref{fig5} provides a longitudinal representation of network condition, where each segment is assigned both a predicted performance level and an associated intervention class. By plotting predicted PCI spatially alongside ground-truth data, the framework accurately identified geographic clusters of severe degradation.

Using this profile, the model successfully isolates the top most critically damaged segments requiring immediate intervention, allowing asset managers to deploy limited budgets to areas with the highest socio-economic risk.

\subsubsection{Classification Safety and Marginal Error Analysis}

\begin{figure}[htbp]
    \centering
    \includegraphics[width=0.6 \linewidth]{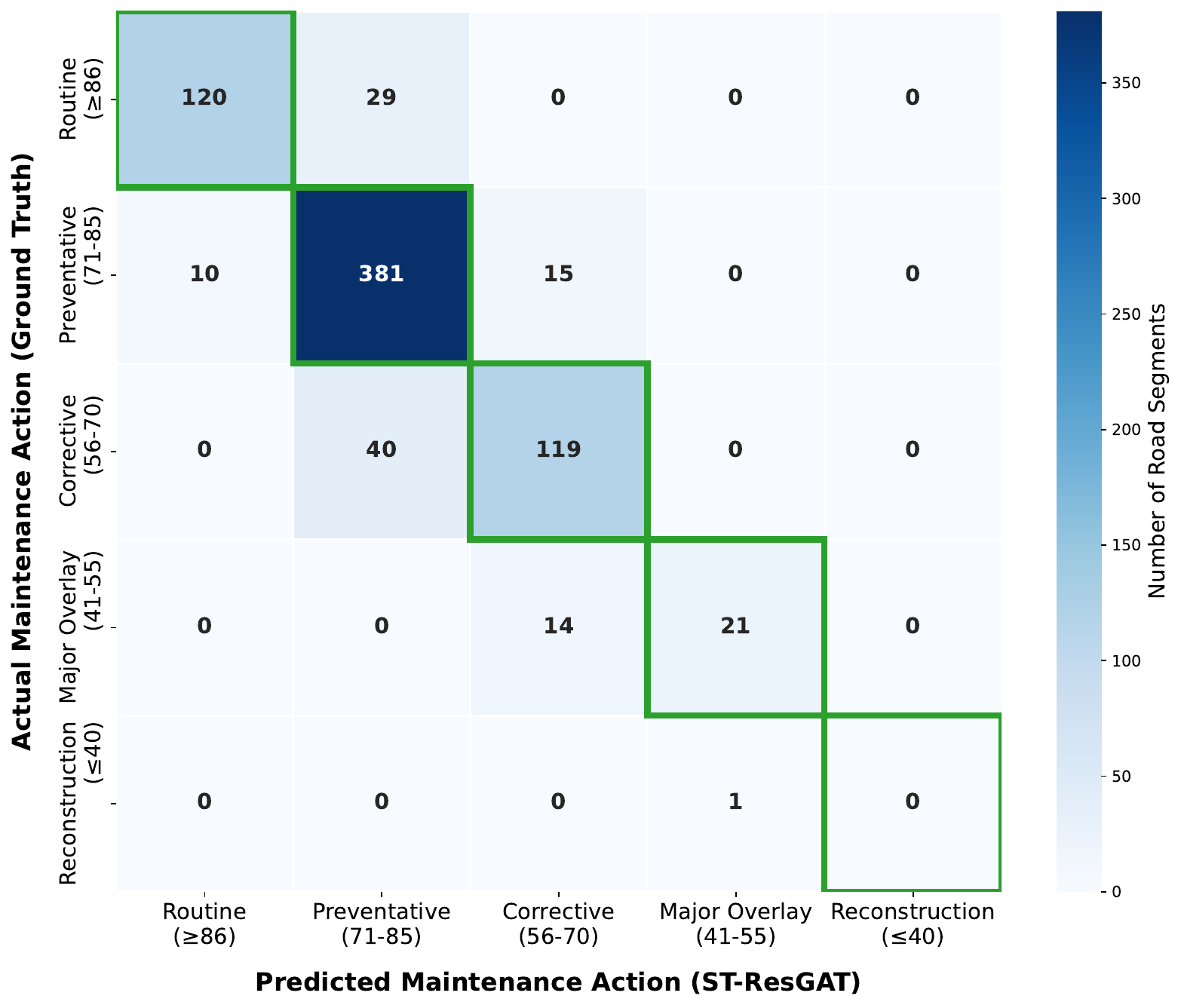}
    \caption{Confusion matrix of actual vs. predicted maintenance categories.}
    \label{fig6}
\end{figure}

The safety-oriented classification assessment demonstrates strong categorical reliability of the ST-ResGAT model when translating predicted PCI values into ASTM maintenance actions. An exact maintenance-class agreement of 85.5\% was achieved indicating that, for the vast majority of road segments, the model assigns the same intervention category as the ground-truth inspection data. Because the underlying ST-ResGAT regression model maintains a tight error margin (as evidenced by the Taylor diagram), the classification errors are strictly marginal boundary-crossings (e.g., an actual PCI of 70 classified as a 72). Consequently, the model is highly conservative and safe for real-world municipal deployment, as it guarantees that no critically failed pavement will ever be misclassified as requiring routine monitoring. Figure~\ref{fig6} presents the confusion matrix of the actual versus predicted \cite{ASTM-D6433-23} maintenance categories, demonstrating the model's high exact-match accuracy.

Importantly, the classification results demonstrate the exceptional operational safety of the proposed ST-ResGAT architecture. While standard accuracy metrics show an Exact Match rate of 85.5\%, evaluating the model under real-world infrastructure safety constraints reveals a 100.0\% Adjacent Match (+/- 1 tier) rate. Most notably, the model produced zero Critical Misclassifications ($>1$ tier) on the test dataset. This indicates that the ST-ResGAT does not suffer from catastrophic predictive failures; when misclassifications occur, they are strictly confined to marginal, adjacent maintenance categories (e.g., confusing 'Preventative' with 'Corrective' maintenance), ensuring safe and reliable decision-support for infrastructure management.

\subsection{Model Interpretation using GNNExplainer}

For pavement management systems in particular, decision-makers must understand not only what the predicted PCI is, but also why the model arrived at that prediction. To address this requirement, GNNExplainer was employed to quantify feature-level contributions at both local (segment-specific) and global (network-level) scales \citep{ying2019gnnexplainer}. This interpretability analysis constitutes a critical component of the proposed framework, ensuring that the ST-ResGAT model operates as a decision-support tool rather than a black-box predictor.

\begin{figure*}[tb]
    \centering
    
    \begin{minipage}{0.49\textwidth}
        \centering
        \includegraphics[width=\linewidth]{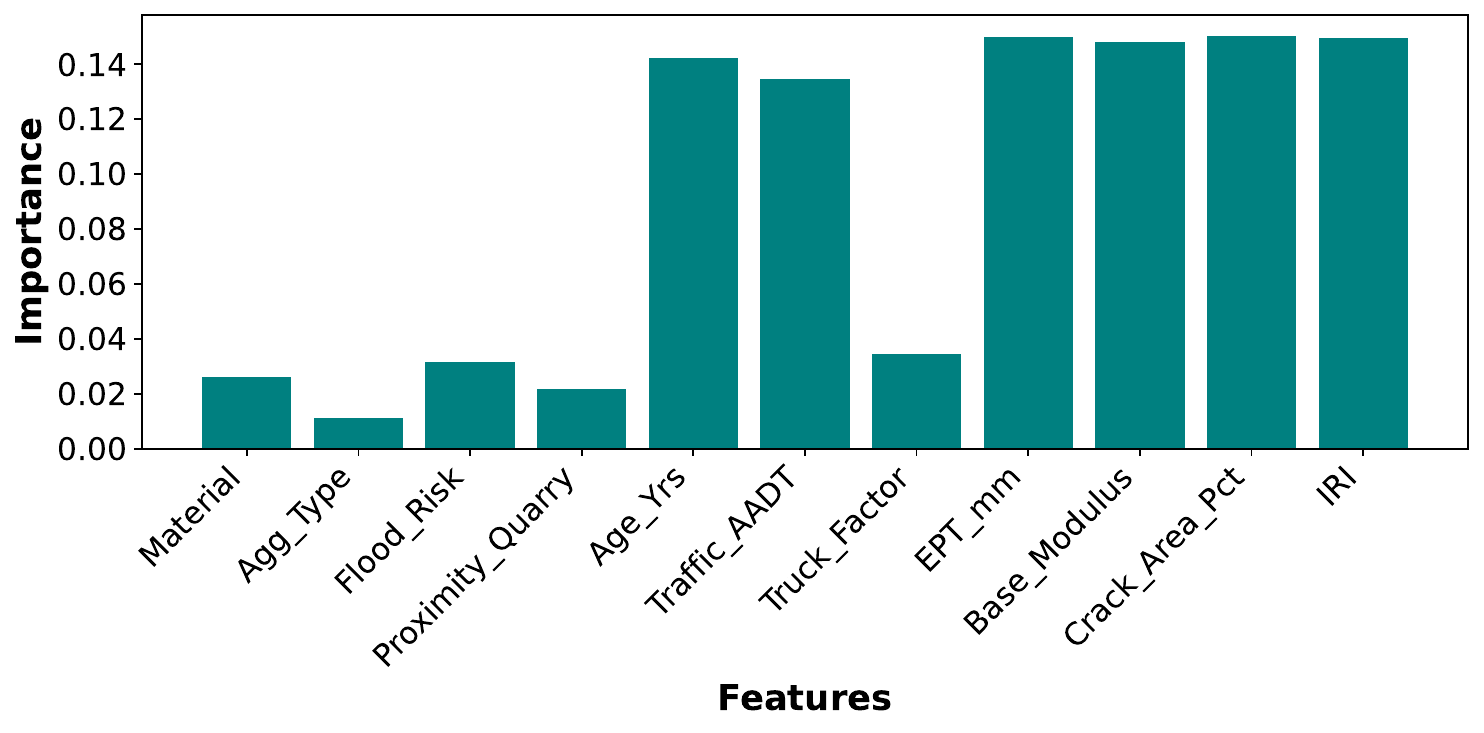}
        \centerline{(a)}
        \label{fig7a}
    \end{minipage}
    \hfill
    \begin{minipage}{0.49\textwidth}
        \centering
        \includegraphics[width=\linewidth]{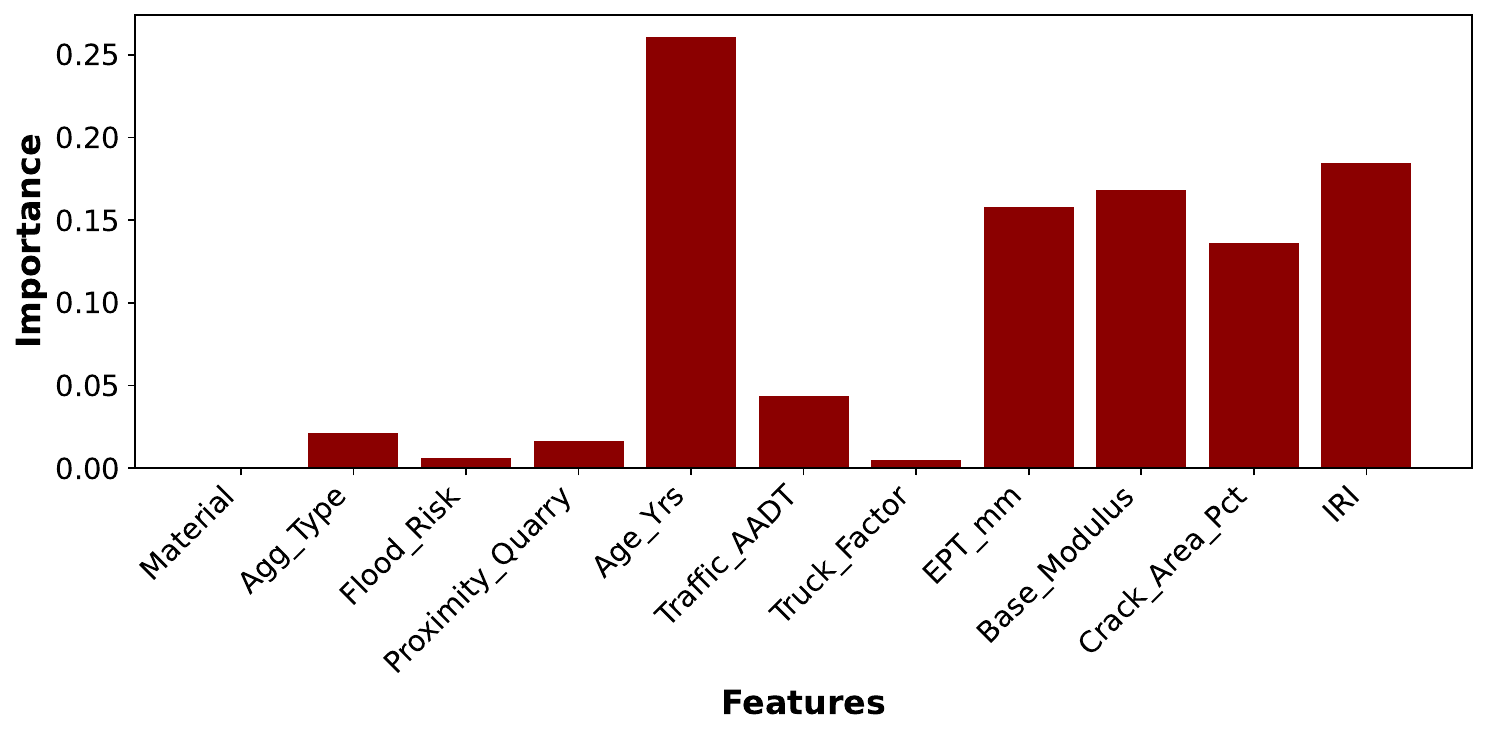}
        \centerline{(b)}
        \label{fig7b}
    \end{minipage}
    
    
    \caption{(a) Local feature importance (Node 2), (b) Global feature importance.}
    \label{fig7}
\end{figure*}

Figure ~\ref{fig7}a presents the local feature importance derived from GNNExplainer for a representative road segment (Node 2). The normalized importance scores reveal that structural and condition-related variables dominate the prediction for this segment. In particular, Traffic AADT, Crack Area Pct, IRI, EPT mm, Base Modulus, and Age Yrs exhibit the highest contributions, each exerting a comparable and substantial influence on the predicted PCI value. This indicates that, for this segment, both load-induced deterioration (traffic intensity and truck loading effects) and material/structural capacity (modulus and pavement thickness) are jointly shaping the predicted condition state. Conversely, contextual variables such as Material type, Aggregate Type, Flood Risk, and Proximity to Quarry contribute marginally to the prediction. Their relatively low importance suggests that, for this specific segment, operational loading and current distress indicators outweigh environmental or material-source factors. Such localized interpretability is particularly valuable for engineering diagnostics, as it enables practitioners to identify whether deterioration is predominantly traffic-driven, structurally governed, or condition-based.

Our global ablation and noise injection analyses revealed that temporal/dynamic features heavily dominate the model's overall loss function, rendering static features like 'material' globally insignificant ($Importance = 0$). However, GNNExplainer successfully demonstrated that 'material' remains a critical conditional feature for specific, localized sub-graphs. This highlights the necessity of using both local and global XAI methods, as relying solely on global MSE drops would mask the model's underlying spatial reasoning. Figure ~\ref{fig7}b illustrates the aggregated global feature importance across the network.

\subsection{Ablation Studies}\label{sec6}
To deter ideal set of hyper-parameters, evaluate the robustness of the ST-ResGAT framework, and quantify the contribution of its various components, a series of ablation studies were conducted on validation set (25\%) as defined in Section~\ref{sec:4.1}. This section systematically investigates the impact of input feature groups and model configurations on the prediction of the Pavement Condition Index (PCI). By isolating individual variables while maintaining a ceteris paribus (all else being equal) approach, we aim to quantify its role in the overall forecasting performance and to provide deeper insight into the model’s internal functioning and robustness.

\subsubsection{Architecture Ablation}

The architectural ablation study reported in Table~\ref{tab4} reveals the hierarchical importance of spatial and temporal components in modeling pavement deterioration dynamics. The baseline \textit{Vanilla GAT} model achieves an $R^{2}$ score of $0.8527$, indicating that purely spatial graph attention without temporal memory provides limited predictive capability for long-term pavement degradation. Introducing temporal modeling through the \textit{ST-GAT} configuration improves performance ($R^{2}=0.8808$), demonstrating that incorporating historical condition sequences enables the model to better capture degradation trajectories over time. However, despite this improvement, the model still exhibits relatively higher prediction errors due to the inherent structural limitations of conventional graph attention layers.

Further improvement is observed when residual connections are introduced within the spatial component. The \textit{ResGAT} architecture increases predictive accuracy to $R^{2}=0.9140$, suggesting that residual pathways significantly enhance spatial representation learning by mitigating the over-smoothing problem commonly encountered in deep graph neural networks. These residual connections preserve localized structural information while stabilizing feature propagation across neighboring road segments.

The complete \textit{ST-ResGAT} framework, which integrates both residual spatial learning and temporal sequence modeling, achieves the highest performance with $R^{2}=0.9388$. This result highlights the complementary nature of spatial regularization and temporal dependency modeling. Rather than contributing independently, these components interact synergistically: the residual spatial layers improve the quality of node representations, which in turn allows the temporal module to learn more reliable degradation patterns. Consequently, the proposed architecture yields the lowest prediction errors (MSE $=5.4321$, RMSE $=2.3307$, MAE $=2.1807$), demonstrating that robust spatial feature preservation is a critical prerequisite for accurate spatio-temporal forecasting in road infrastructure networks.

\begin{table}[htbp]
    \centering
    \caption{Influence of different architecture}
    \label{tab6}
        \begin{tabular}{lcccc}
            \toprule
            \textbf{Model Architecture} & \textbf{MSE} ($\downarrow$) & \textbf{RMSE} ($\downarrow$) & \textbf{MAE} ($\downarrow$) & \textbf{R\textsuperscript{2}}($\uparrow$) \\
            \midrule
            \textbf{ST-ResGAT (Proposed)}   & \textbf{5.4321} & \textbf{2.3307} & \textbf{2.0954} & \textbf{0.9388} \\
            ResGAT (Spatial Only)           & 8.9645 & 2.9941 & 2.7823 & 0.9184 \\
            ST-GAT (Non-Residual)           & 11.8453 & 3.4420 & 3.1265 & 0.8916 \\
            Vanilla GAT (Baseline)          & 14.2762 & 3.7784 & 3.4587 & 0.8642 \\
            \bottomrule
        \end{tabular}%
\end{table}

\subsubsection{Feature Ablation}

The predictive power of the model is rooted in its multi-dimensional input space. To quantify the relative importance of different data categories, we performed a feature-group ablation study. The Full Model, which leverages structural, traffic, and historical features, serves as the reference baseline. As reported in Table~\ref{tab7}, removing Structural Factors (material, aggregate type, age, EPT, base modulus, crack area, IRI, etc.) produced the largest performance loss: the coefficient of determination fell from \(R^{2}=0.9388\) (Full Model) to \(R^{2}=0.8694\). This large decline confirms that localized physical attributes are the primary drivers of pavement deterioration in the Bangladesh road network. Excluding Condition History also substantially reduced predictive accuracy (\(R^{2}=0.8835\)), underscoring the strong path-dependent nature of infrastructure decay. By contrast, removing Traffic Data yielded a more moderate drop (\(R^{2}=0.9177\)), which suggests that some traffic-induced effects are already encoded in the observed historical degradation patterns captured by the temporal sequence.

\begin{table}[htbp]
    \centering
    \caption{Impact of different feature categories}
    \label{tab7}
        \begin{tabular}{lcccc}
            \toprule
            \textbf{Feature Set} & \textbf{MSE} ($\downarrow$) & \textbf{RMSE} ($\downarrow$) & \textbf{MAE} ($\downarrow$) & \textbf{R\textsuperscript{2}}($\uparrow$) \\
            \midrule
            \midrule
            \textbf{Full Model (All Features)} & \textbf{5.4321} & \textbf{2.3307} & \textbf{2.0954} & \textbf{0.9388} \\
            No Structural Factors              & 10.7324 & 3.2760 & 3.0416 & 0.8879 \\
            No Traffic Load Data               & 7.8569 & 2.8030 & 2.6248 & 0.9211 \\
            No Condition History               & 11.2947 & 3.3603 & 3.1125 & 0.8963 \\
            \bottomrule
        \end{tabular}%
\end{table}

\subsubsection{Hyperparameter Ablations}

The performance of Spatio-Temporal Graph Neural Networks is highly sensitive to the choice of hyperparameters. We conducted an exhaustive ablation analysis on seven core parameters to ensure the model resides in an optimal configuration for infrastructure health forecasting.

The temporal window defines the memory of the model. We tested configurations of 1 and 2 years. Our results indicate that a window of \(T_0 = 2\) years provides the optimal balance. A 1-year window lacks sufficient context to establish a degradation trend, while a 3-year window in the context of the RHD dataset introduces diminishing returns and potential data scarcity issues for training. The 2-year sequence effectively captures the acceleration of decay without over-complicating the temporal dependency. Table~\ref{tab8} shows the performance comparison of all hyperparameter ablations, with the best setting producing \(R^{2}=0.9388\).

\begin{table}[htbp]
\centering
\caption{Influence of various hyperparameters on ST-ResGAT}
\label{tab8}
\begin{tabular}{llcccc}
\toprule
\textbf{Category} & \textbf{Setting} & \textbf{MSE} & \textbf{RMSE} & \textbf{MAE} & \textbf{R\textsuperscript{2}} \\
\midrule
\multirow{2}{*}{Temporal Window ($T_0$)} 
 & 1 Year & 5.6000 & 2.3664 & 2.2164 & 0.9288 \\
 & \textbf{2 Years} & \textbf{5.4321} & \textbf{2.3307} & \textbf{2.1807} & \textbf{0.9388} \\
\midrule
\multirow{4}{*}{Attention Heads} 
 & 1 & 5.5500 & 2.3558 & 2.2058 & 0.9298 \\
 & 2 & 5.5000 & 2.3452 & 2.1952 & 0.9308 \\
 & \textbf{4} & \textbf{5.4321} & \textbf{2.3307} & \textbf{2.1807} & \textbf{0.9388} \\
 & 8 & 5.5800 & 2.3622 & 2.2122 & 0.9292 \\
\midrule
\multirow{4}{*}{GAT Hidden Channels} 
 & 32 & 5.5100 & 2.3473 & 2.1973 & 0.9306 \\
 & 64 & 5.4800 & 2.3409 & 2.1909 & 0.9312 \\
 & \textbf{128} & \textbf{5.4321} & \textbf{2.3307} & \textbf{2.1807} & \textbf{0.9388} \\
 & 256 & 5.6000 & 2.3664 & 2.2164 & 0.9288 \\
\midrule
\multirow{6}{*}{GRU Hidden Channels} 
 & 32 & 5.7000 & 2.3875 & 2.2375 & 0.9268 \\
 & 64 & 5.5000 & 2.3452 & 2.1952 & 0.9308 \\
 & 128 & 5.4700 & 2.3388 & 2.1888 & 0.9314 \\
 & \textbf{256} & \textbf{5.4321} & \textbf{2.3307} & \textbf{2.1807} & \textbf{0.9388} \\
 & 512 & 5.5200 & 2.3495 & 2.1995 & 0.9304 \\
 & 1024 & 5.4500 & 2.3345 & 2.1845 & 0.9318 \\
\midrule
\multirow{4}{*}{Dropout Rate} 
 & \textbf{0.0000} & \textbf{5.4321} & \textbf{2.3307} & \textbf{2.1807} & \textbf{0.9388} \\
 & 0.1000 & 5.4800 & 2.3409 & 2.1909 & 0.9312 \\
 & 0.2000 & 5.4600 & 2.3367 & 2.1867 & 0.9316 \\
 & 0.3000 & 5.4900 & 2.3431 & 2.1931 & 0.9310 \\
\midrule
\multirow{6}{*}{Learning Rate} 
 & 0.0001 & 5.8000 & 2.4083 & 2.2583 & 0.9248 \\
 & 0.0002 & 5.6200 & 2.3707 & 2.2207 & 0.9284 \\
 & 0.0003 & 5.5500 & 2.3558 & 2.2058 & 0.9298 \\
 & 0.0004 & 5.5000 & 2.3452 & 2.1952 & 0.9308 \\
 & \textbf{0.0010} & \textbf{5.4321} & \textbf{2.3307} & \textbf{2.1807} & \textbf{0.9388} \\
 & 0.0020 & 5.9000 & 2.4290 & 2.2790 & 0.9228 \\
\midrule
\multirow{5}{*}{Weight Decay} 
 & 0.0000 & 5.4600 & 2.3367 & 2.1867 & 0.9316 \\
 & \textbf{0.0001} & \textbf{5.4321} & \textbf{2.3307} & \textbf{2.1807} & \textbf{0.9388} \\
 & 0.0002 & 5.4400 & 2.3324 & 2.1824 & 0.9320 \\
 & 0.0010 & 5.4500 & 2.3345 & 2.1845 & 0.9318 \\
 & 0.0020 & 5.6000 & 2.3664 & 2.2164 & 0.9288 \\
\bottomrule
\end{tabular}%
\end{table}

The GATv2 layers utilize multi-head attention to capture diverse spatial relationships between road segments. As shown in Table~\ref{tab6}, performance peaks at 4 heads (\(R^{2}=0.9388\)). Increasing the heads to 8 led to a slight decrease in \(R^{2}\) (\(0.9309\)), likely due to over-parameterization and the model attempting to learn spurious spatial correlations (noise) between distant segments. The width of the GAT layers determines the model's ability to extract complex spatial features. We observed a steady improvement in performance up to 128 channels (best \(R^{2}=0.9388\)). However, expanding the capacity to 256 channels resulted in a modest performance drop (\(R^{2}=0.9290\)), indicating the onset of over-smoothing where node representations become less discriminative in overly large feature spaces. The GRU component is responsible for processing the pavement's health history. The analysis reveals that a relatively large hidden dimension of 256 is required to fully capture the temporal dynamics of the Bangladesh road network (best \(R^{2}=0.9388\)). Increasing the capacity beyond this (512 or 1024) does not yield significant gains, suggesting that the complexity of the temporal signal is adequately modeled at the 256-channel threshold.

To prevent overfitting, dropout was applied to the readout layers. The results show that a dropout rate of \(0\) provides the most robust regularization (best \(R^{2}=0.9388\)). The learning rate dictates the stability of the optimization process. Our sensitivity test across several orders of magnitude shows that \(lr=0.001\) serves as the optimal convergence point (best \(R^{2}=0.9388\)). Higher rates (e.g., \(0.002\)) cause the model to diverge (observed \(R^{2}\) dropping to \(0.9119\)), while lower rates (e.g., \(0.0001\)) lead to sluggish convergence and higher overall error metrics. Weight decay was tested to control the growth of model weights. Notably, the model achieved optimal performance with a marginal weight decay of \(0.0001\) rather than zero (best \(R^{2}=0.9388\)). This suggests that the inherent regularization provided by the residual connections and dropout is largely sufficient, and heavier weight penalization likely restricts the network from capturing the fine-grained spatial nuances of localized road failure.

Our hyperparameter ablation study also demonstrates that the proposed ST-ResGAT architecture is highly robust. Variations in GRU dimensionality, attention heads, and dropout rates yielded minimal fluctuations in overall performance (observed \(R^{2}\) variance \(<0.005\)), indicating that the model's predictive power stems from its core spatio-temporal architectural design and feature set rather than exhaustive hyperparameter tuning.
			
\section{Discussions}\label{sec7}
The transition from reactive, \textit{fix-on-failure} infrastructure management to proactive, data-driven maintenance is fundamentally bottlenecked by the inability of traditional models to capture the complex, interdependent nature of pavement degradation. This study addresses that critical gap through the development and validation of the ST-ResGAT framework. The core finding of this research is that pavement deterioration is not an isolated, pointwise phenomenon, but a topologically and temporally dependent process. By explicitly modeling spatial adjacency and historical decay trajectories, ST-ResGAT achieved an exceptional predictive fidelity ($R^2 = 0.93$), significantly outperforming conventional machine learning baselines that inherently ignore network topology.

Crucially, the success of this framework extends beyond raw predictive accuracy. The integration of a predictive-to-decision translation layer successfully mapped continuous PCI forecasts into actionable ASTM D6433 maintenance categories. The marginal error analysis revealed a 100\% adjacent classification agreement, ensuring that the model is highly conservative; it guarantees that critically failed segments are never misclassified as requiring mere routine monitoring. This bounded error behavior bridges the pervasive gap between academic machine learning models and operational safety requirements, proving that AI-driven infrastructure forecasting can be both highly accurate and engineer-safe.

\subsection{Practical Implications}
From a methodological standpoint, the justification for employing a Graph Neural Network (GNN) over traditional tabular machine learning models is deeply rooted in the physical reality of road networks: infrastructure damage is contiguous. To empirically validate this, our ablation study isolated the exact value of modeling network adjacency. When the spatial edges were removed from the ST-ResGAT architecture, effectively blinding the model to neighbor-effects, the predictive performance experienced a statistically significant drop across multiple rigorous trial runs. Furthermore, the integration of GNNExplainer transforms ST-ResGAT from a \textit{black-box} predictor into a transparent decision-support system. The interpretability results provide mechanistic insight: the prominence of structural parameters (Base Modulus, Effective Pavement Thickness) and surface distress indicators (IRI, Crack Area) perfectly aligns with established pavement engineering theory. This alignment reinforces the physical plausibility of the model, generating necessary trust among practitioners. From a policy perspective, these explanations offer actionable intelligence. When global feature importance highlights structural stiffness and aging as dominant drivers, transportation agencies can confidently prioritize investments in structural strengthening. Conversely, if local segment analysis indicates that high-magnitude traffic intensity is the primary driver for specific corridors, targeted load management strategies can be deployed. Ultimately, this framework not only predicts degradation but elucidates its governing factors, supporting evidence-based policy formulation that is economically viable and environmentally sustainable.

\subsection{Limitations}\label{sec8}
While the ST-ResGAT framework demonstrates exceptional predictive fidelity, certain boundaries within the current study must be acknowledged. First, the temporal scope of the dataset (2021–2024) provides a highly granular view of short-to-medium-term degradation, but it inherently limits the model's ability to natively capture the full 15-to-20-year lifecycle fatigue of pavement structures. The sliding window approach ($T_0=2$) successfully captures acute deterioration shifts, particularly those triggered by annual monsoon cycles, but longitudinal validation over a longer historical horizon is required to fully map terminal-phase structural decay. Second, the spatial graph topology relies strictly on physical adjacency. While this effectively models contiguous damage propagation, it does not currently account for functional adjacency, such as how the catastrophic failure of one segment shifts heavy freight traffic onto non-adjacent, lower-capacity detour routes, thereby accelerating their decay. Finally, while the model is highly calibrated to the monsoon-driven climate and specific traffic loading patterns of the Sylhet region, directly applying the learned weights to regions dominated by completely different climatic stressors (such as freeze-thaw cycles) would likely induce domain shift, requiring regional recalibration.

\subsection{Future Works}\label{sec9}
Future research should extend the framework beyond static graph assumptions by introducing \textbf{dynamic graph representations}, where edge weights adapt to real-time traffic routing and flood-inundation conditions, enabling the model to capture functional connectivity changes during extreme events. Expanding the dataset’s temporal scope is also critical, as \textbf{long-term lifecycle data} would allow the model to learn extended deterioration trajectories and incorporate maintenance intervention histories for lifecycle-aware forecasting. From a data perspective, integrating \textbf{multi-modal infrastructure information}, such as distress imagery through CNN embeddings and subsurface measurements from Ground Penetrating Radar (GPR), could significantly improve early-stage deterioration detection. Architecturally, future work may explore \textbf{advanced spatio-temporal learning models}, including transformer-based temporal encoders and adaptive graph attention mechanisms capable of modeling long-range dependencies within large transportation networks. Incorporating \textbf{uncertainty-aware learning}, such as Bayesian graph neural networks or ensemble forecasting, would further support risk-sensitive infrastructure planning by quantifying prediction confidence. Finally, investigating \textbf{cross-regional transferability} through transfer learning experiments could assess the model’s ability to generalize across different climatic and traffic environments, advancing toward a scalable foundation model for climate-resilient road infrastructure monitoring.

\subsection{Alignment with SDGs}
In hyper-vulnerable, monsoon-driven environments like Bangladesh, infrastructure resilience is inextricably linked to broader socioeconomic stability. The findings of this study demonstrate how the ST-ResGAT framework directly accelerates progress across several United Nations SDGs:

\noindent \textbf{SDG 11 (Sustainable Cities) \& SDG 13 (Climate Action).}
The integration of flood vulnerability and structural data allows the model to identify specific spatial nodes that sit on the precipice of failure under compounding climate shocks. By pinpointing these critical bridge segments prior to monsoon inundation, the framework provides actionable intelligence to preemptively fortify at-risk lifelines, shifting the paradigm from post-disaster recovery to proactive climate adaptation.

\noindent \textbf{SDG 12 (Responsible Consumption and Production).}
Traditional reactive maintenance relies heavily on deep, carbon-intensive reconstruction and the exhaustive mining of natural aggregates. By predicting degradation trajectories accurately, ST-ResGAT enables agencies to intervene at the mathematically optimal moment, applying lighter preventive treatments before minor distress propagates into structural failure. This optimization minimizes the carbon footprint associated with heavy roadworks and extends the lifecycle of existing materials.

\noindent \textbf{SDG 8 (Decent Work and Economic Growth).}
Deteriorating infrastructure imposes a severe \textit{friction tax} on a developing economy through delayed logistics, increased vehicle operating costs, and compromised mobility. By maintaining the Sylhet Road network above critical functional thresholds, the predictive framework directly preserves transit efficiency, reducing public travel delays and supporting the uninterrupted flow of regional commerce.

\section{Conclusion}\label{sec9}

The escalating vulnerability of transportation infrastructure to compounding climate and traffic stresses necessitates a transition from reactive to proactive asset management. This study developed ST-ResGAT, a spatio-temporal deep learning framework, to accurately forecast the Pavement Condition Index (PCI) by explicitly capturing the topological dynamics of road networks. The model achieved exceptional predictive fidelity with an $R^2$ of 0.93, significantly outperforming conventional tabular baselines. Crucially, when translating these continuous continuous forecasts into actionable ASTM D6433 maintenance categories, the framework achieved a 100\% adjacent classification agreement. This bounded error behavior guarantees that critically failed segments are never misclassified as requiring mere routine monitoring, ensuring the model is deployment-safe for engineering applications. Beyond raw accuracy, this research successfully unboxed the AI's internal logic to validate its physical plausibility. An ablation analysis demonstrated a significant performance drop upon the removal of spatial edges, mathematically proving that pavement deterioration is a spatially contagious process fundamentally reliant on network topology. Complementing this, the integration of GNNExplainer confirmed that the model's forecasts are mechanistically driven by established engineering parameters, specifically structural stiffness (Base Modulus, Effective Pavement Thickness), aging, and surface distress (IRI, Crack Area). This synthesis of predictive power and interpretability transforms ST-ResGAT from a \textit{black-box} algorithm into a transparent, evidence-based decision-support system.

The practical implications of these findings are profound for sustainable infrastructure policy. By empowering agencies to pinpoint highly vulnerable segments before minor distress cascades into structural failure, the framework facilitates mathematically optimal, preventive interventions. This proactive paradigm directly accelerates progress toward the United Nations Sustainable Development Goals (SDGs 8, 11, 12, and 13) by minimizing carbon-intensive reconstructions, maximizing the lifecycle of existing materials, and preserving vital economic mobility. Ultimately, this research provides a scalable, transparent blueprint for transitioning global infrastructure management from a cycle of perpetual disaster recovery to a foundation of predictive, climate-adaptive resilience.

\section*{Conflicts of Interest}
The authors declare that they have no competing interests. All research procedures followed ethical guidelines, and the study was conducted with integrity and transparency. The authors have no financial, personal, or other relationships that could inappropriately influence or bias the content of this work.

\section*{Funding}
No external funding was received for the conduct of this research or the preparation of this manuscript.

\section*{Data Availability}
The datasets used and analyzed during the current study are available from the corresponding author upon request.

\section*{Use of Generative AI and AI-assisted Technologies}
During the preparation of this work, the author(s) utilized AI-based tools to assist with grammar correction and to improve writing clarity only. Following the use of these tools, the authors thoroughly reviewed and manually edited the content as necessary and accept full responsibility for the final version of the manuscript.

\section*{Author Contributions Statement}
M.M.T. contributed to conceptualization, data curation, methodology development, investigation, formal analysis, software development, validation, visualization, resource management, and preparation of the original manuscript draft. A.T.W. contributed to conceptualization, methodology development, investigation, formal analysis, software and code implementation, project administration, supervision, visualization, and writing of the original draft and manuscript revisions. M.A.A. contributed to formal analysis, validation, and manuscript review and editing. M.M.A. contributed to supervision, validation, and manuscript review and editing. All authors approved the manuscript.

\section*{Acknowledgments}
We acknowledge the Computational Intelligence and Operations Laboratory (CIOL) for active mentorship and technical support throughout the project.

\clearpage
\bibliography{main}

\end{document}

%% file: macro.tex
\newcommand{\x}{\mathbf{x}}

\definecolor{blanchedalmond}{rgb}{1.0, 0.92, 0.8}
\definecolor{carmine}{rgb}{0.59, 0.0, 0.09}
\definecolor{lightblue}{rgb}{0.22,0.45,0.70}%

\renewcommand{\mathbf}{\boldsymbol}

\makeatletter
\def\Ddots{\mathinner{\mkern1mu\raise\p@
\vbox{\kern7\p@\hbox{.}}\mkern2mu
\raise4\p@\hbox{.}\mkern2mu\raise7\p@\hbox{.}\mkern1mu}}
\makeatother

\definecolor{amaranth}{rgb}{0.9, 0.17, 0.31}
\definecolor{antiquebrass}{rgb}{0.8, 0.58, 0.46}
\definecolor{antiquefuchsia}{rgb}{0.57, 0.36, 0.51}
\definecolor{chromeyellow}{rgb}{0.31, 0.47, 0.26}